\documentclass[conference]{IEEEtran}

\usepackage[available,functional]{ndssbadges}

\usepackage{amsmath,amssymb,amsfonts}
\newtheorem{definition}{Definition}

\usepackage{bm}
\usepackage{cite}
\usepackage{bbding}
\usepackage{wasysym}
\usepackage{algorithm}
\usepackage[noend]{algpseudocode}
\usepackage{multirow}
\usepackage{mathrsfs}
\usepackage{balance}
 
\usepackage{url}
\usepackage{tikz}
\usepackage{savesym}
\savesymbol{iint}
\savesymbol{iiint}
\usepackage{amsmath}

\usepackage{hyperref}
\hypersetup{
	colorlinks=true,
	linkcolor=red,
	filecolor=magenta,      
	urlcolor=black,
	citecolor=black
}

\usepackage[numbers, sort&compress]{natbib}

\usepackage{subcaption}

\usepackage{tabu}
\usepackage{threeparttable}
\usepackage{tabularx}
\usepackage{tikz}
\newcommand*\emptycirc[1][1ex]{\tikz\draw[thick] (0,0) circle (#1);} 
\newcommand*\halfcirc[1][1ex]{%
  \begin{tikzpicture}
  \draw[fill] (0,0)-- (90:#1) arc (90:270:#1) -- cycle ;
  \draw[thick] (0,0) circle (#1);
  \end{tikzpicture}}
\newcommand*\fullcirc[1][1ex]{\tikz\fill (0,0) circle (#1);}

\newcommand{\ec}{\emptycirc[0.8ex]}
\newcommand{\hc}{\halfcirc[0.8ex]}
\newcommand{\fc}{\fullcirc[0.9ex]}

\usepackage{bbding}

\usepackage{filecontents}

\usepackage{bbding}
\usepackage{wasysym}
\usepackage{booktabs}
\usepackage{makecell}

\usepackage{bbm}

\usepackage{color}
\newcommand{\mr}{\textcolor{black}}

\usepackage{xurl}

\def\BibTeX{{\rm B\kern-.05em{\sc i\kern-.025em b}\kern-.08em
    T\kern-.1667em\lower.7ex\hbox{E}\kern-.125emX}}
\begin{document}


\title{GraphGuard: Detecting and Counteracting \\ Training Data Misuse in Graph Neural Networks}

\author{\IEEEauthorblockN{Bang Wu\IEEEauthorrefmark{1}\IEEEauthorrefmark{2}\textsuperscript{1},
He Zhang\IEEEauthorrefmark{2},
Xiangwen Yang\IEEEauthorrefmark{2}, 
Shuo Wang\IEEEauthorrefmark{1}\IEEEauthorrefmark{3},
Minhui Xue\IEEEauthorrefmark{1},
Shirui Pan\IEEEauthorrefmark{4} and
Xingliang Yuan\IEEEauthorrefmark{2}}
\IEEEauthorblockA{\IEEEauthorrefmark{1}CSIRO's Data61, Australia}
\IEEEauthorblockA{\IEEEauthorrefmark{2}Monash University, Australia}
\IEEEauthorblockA{\IEEEauthorrefmark{3}Shanghai Jiao Tong University, China}
\IEEEauthorblockA{\IEEEauthorrefmark{4}Griffith University, Australia}}

\IEEEoverridecommandlockouts
\makeatletter\def\@IEEEpubidpullup{6.5\baselineskip}\makeatother
\IEEEpubid{\parbox{\columnwidth}{
    Network and Distributed Systems Security (NDSS) Symposium 2024\\
    26 February - 1 March 2024, San Diego, CA, USA\\
    ISBN 1-891562-93-2\\
    https://dx.doi.org/10.14722/ndss.2024.24441\\
    www.ndss-symposium.org
}
\hspace{\columnsep}\makebox[\columnwidth]{}}

\maketitle

\begin{abstract}
The emergence of Graph Neural Networks (GNNs) in graph data analysis and their deployment on Machine Learning as a Service platforms have raised critical concerns about data misuse during model training. This situation is further exacerbated due to the lack of transparency in local training processes, potentially leading to the unauthorized accumulation of large volumes of graph data, thereby infringing on the intellectual property rights of data owners.
Existing methodologies often address either data misuse detection or mitigation, and are primarily designed for local GNN models rather than cloud-based MLaaS platforms. These limitations call for an effective and comprehensive solution that detects and mitigates data misuse without requiring exact training data while respecting the proprietary nature of such data.
This paper introduces a pioneering approach called GraphGuard, to tackle these challenges. We propose a training-data-free method that not only detects graph data misuse but also mitigates its impact via targeted unlearning, all without relying on the original training data. Our innovative misuse detection technique employs membership inference with radioactive data, enhancing the distinguishability between member and non-member data distributions. For mitigation, we utilize synthetic graphs that emulate the characteristics previously learned by the target model, enabling effective unlearning even in the absence of exact graph data. 
We conduct comprehensive experiments utilizing four real-world graph datasets to demonstrate the efficacy of GraphGuard in both detection and unlearning. 
We show that GraphGuard attains a near-perfect detection rate of approximately $100\%$ across these datasets with various GNN models. 
In addition, it performs unlearning by eliminating the impact of the unlearned graph with a marginal decrease in accuracy (less than $5\%$).
\end{abstract}

\section{Introduction}



Graph Neural Networks (GNNs) have gained prominence in graph data analysis due to their exceptional performance and wide-ranging applications~\cite{KipfW17,0022WZ20, ZhengZLZWP23, JinWZZDXP23}, including e-Commerce, drug discovery, and protein folding ~\cite{Yang19,JinSEIZCJB21,JumperEPGFRTBZP21,WangHZZZL18,XieCYZL18,ChenLZY18,DengH21,WanLWW21}. 
Currently, Machine Learning as a Service (MLaaS) platforms~\cite{aws-sagemaker,azure-machine-learning,google-vertex-ai,ibm-wml-zos} have revolutionized the deployment of GNN models by enabling model developers \cite{LiuWYY22} to deploy them in the cloud and providing public APIs to end users. 
However, MLaaS platforms face transparency challenges, especially concerning models developed and trained locally. 
The lack of visibility in the local training process complicates monitoring for irregularities or illicit activities, raising concerns about potential \textit{data misuse} during model training.

\mr{
The \textit{data misuse} problem during ML model training presents significant practical and security concerns. 
As deep learning models intrinsically rely on vast data sets for optimal functionality, model developers have a propensity to gather large amounts of data, sometimes through unauthorized means, occasionally or intentionally. 
Such misuse undermines the intellectual property rights of data owners. 
For example, recent ML tool developers, who deploy their ML models via MLaaS platforms like OpenAI and LensaAI, have been sued by authors and artists for unauthorized utilization of their intellectual content~\cite{Frenkel_Thompson_2023,Mattei_2022}. 
These lawsuits have prompted policy development and legislation, with the EU introducing policies to prevent the misuse of training data~\cite{LLP_2023}. 
}

\mr{
The data misuse issue also gains prominence in the context of GNNs since they are utilized in sensitive domain applications, such as using GNNs to predict properties in Protein-Protein Interaction (PPI) graphs in pharmaceutical research~\cite{mitchell2018artificial,kop2019ai,oliva2022management,kooli2022artificial}. 
In such scenarios, nodes signify proteins and edges indicate their interactions~\cite{LiuYCJ20,wang2023assessment}. 
The construction of these graphs demands extensive experimentation and significant financial investments, making them valuable intellectual property~\cite{melo2016machine,mrowka2001there}. 
Many pharmaceutical firms, as GNN model developers, are seeking to leverage GNN to facilitate drug discovery, disease prevention, and diagnosis~\cite{YouZLL17,melo2016machine,yang2019analyzing}. 
These companies might exploit data without owners' consent, seeking commercial gains through MLaaS predictions, thus bypassing the costs of data acquisition. 
Such unauthorized use severely breaches the data owner's IP rights. 
Therefore, there is a pressing need to protect graph data against potential data misuse. 
}

\begin{table*}[t]
\centering
\footnotesize
\renewcommand\arraystretch{1.3}
\caption{Comparison between GraphGuard (our design) and Related Works. \ec \ indicates ``Not Considered", \hc \ indicates ``Partially Considered", and \fc \ indicates ``Fully Considered".}
\label{tab:related_works}
\begin{threeparttable}
\begin{tabularx}{\textwidth}{cc|c|*{4}{>{\centering\arraybackslash}X}}
\hline
\multicolumn{2}{c|}{\multirow{2}{*}{\textbf{Methods}}} &
  \textbf{Scope} &
  \multicolumn{4}{c}{\textbf{Requirements}} \\ \cline{3-7} 
 &
 &
  \multicolumn{1}{c|}{\textbf{GNNs}} &
  \multicolumn{1}{c|}{\textbf{R1 - Detectable}} &
  \multicolumn{1}{c|}{\textbf{R2 - Remedial}} &
  \multicolumn{1}{c|}{\textbf{R3 - Data Privatization}} &
  \multicolumn{1}{c}{\textbf{R4 - Model Agnostic}} \\
\hline
\multirow{2}{*}{Passive MIA} & \cite{CarliniCN0TT22}, \cite{Yuan022}, \cite{0001MMBS22}, \cite{Choquette-ChooT21} & \ec & \fc & \ec & \ec & \fc \\
 & \cite{OlatunjiNK21}, \cite{WuYPY21}, \cite{abs-2102-05429}, \cite{ContiLPX22}  & \fc & \fc & \ec & \ec & \fc \\
\hline
Active MIA & ~\cite{SablayrollesDSJ20}, \cite{TramerSJLJ0C22} & \ec & \fc & \ec & \ec & \hc \\
\hline
\multirow{2}{*}{\begin{tabular}[c]{@{}c@{}}Unlearning-\\ (SISA-based)\end{tabular}} & \cite{BourtouleCCJTZL21} & \ec & \ec & \fc & \ec & \ec \\
 & \cite{Chen000H022} & \fc & \ec & \fc & \ec & \ec \\
\hline
\multirow{2}{*}{\begin{tabular}[c]{@{}c@{}}Unlearning-\\ (Others)\end{tabular}} & \cite{CaoY15}, \cite{WarneckePWR23}, \cite{ChundawatTMK23}, \cite{KimW22}, \cite{Zhang0ZCL22} & \ec & \ec & \fc & \hc & \fc \\
 & \cite{chien2022certified}, \cite{cheng2023gnndelete}, \cite{WuYQS0023}, \cite{0003CM23} & \fc & \ec & \fc & \ec & \hc \\
\hline
\multicolumn{2}{c|}{\textbf{GraphGuard} (Our Design)} & \fc & \fc & \fc & \fc & \fc \\
\hline
\end{tabularx}
\end{threeparttable}
\end{table*}

\begingroup\renewcommand\thefootnote{\textsuperscript{1}}
\footnotetext{This work was partially done when Bang Wu was at Monash University. }
\endgroup

\noindent
\mr{
\textbf{Existing Studies.} 
Existing methods are primarily designed for local deployment rather than cloud deployment of GNN models, rendering them less compatible with MLaaS platforms. 
In addition, they usually focus only on detecting or mitigating data misuse. 
\textit{1) Detecting data misuse. } 
One common approach to determine whether specific graph data were used during model training is membership inference~\cite{abs-2102-05429,WuYPY21,LiuZCLL22,ContiLPX22}. 
However, they typically require analyzing outputs from the target model using original target samples, which might be less practical for detecting data misuse in MLaaS. 
Specifically, data owners have to transfer their exact training samples to the cloud for membership inference, where they may be hesitant to do so due to the sensitivity of their graph data, intellectual property considerations, and data usage agreements. 
\textit{2) Mitigating data misuse. }
Most existing studies concentrate on mitigating data misuse through unlearning~\cite{Chen000H022,chien2022certified,BourtouleCCJTZL21,WuYQS0023}, which involves removing the influence of certain training samples from the trained model. 
However, they call for particular function blocks in their GNN architecture or GNN training process to ensure unlearning, which is incompatible with general GNN models deployed on MLaaS platforms. 
Additionally, they also require the server to store or have access to the training graph for unlearning purposes, making them not feasible given the data owners' intellectual property considerations on their data transfer or storage on the cloud server. 
}

\noindent \textbf{Our Proposal.} 
In this paper, we introduce an integrated pipeline, GraphGuard, to shield graph data from potential misuse in GNNs deployed via MLaaS. 
We start by formalizing the problem of data misuse in GNNs under MLaaS and listing the requirements for effective implementation in the context of MLaaS. 
Next, we propose a comprehensive pipeline that satisfies these requirements, including considerations of \textit{misuse detection} (\textbf{R1}) and \textit{mitigation} (\textbf{R2}), \textit{data privatization} (\textbf{R3}), and \textit{GNN model agnostic} (\textbf{R4}) (refer to Tab.~\ref{tab:related_works}). 
The GraphGuard includes two main modules: proactive misuse detection and training-graph-free unlearning. 
The former safeguards data during detection, eliminating the need to transfer graph structures or model parameters among entities to protect their data privatization. 
The latter achieves mitigation also by considering such data privatization, and adding no extra assumptions on the training GNN model architecture.

\noindent \textbf{Technical Challenges.} 
Meeting all four requirements simultaneously is a non-trivial task. For example, in striving to fulfill both \textbf{R1} and \textbf{R3}, we observe that the efficacy of misuse detection techniques, such as membership inference (\textbf{R1}), could be significantly reduced when avoiding the transmission of graph structures or model parameters among entities (\textbf{R3}). Membership inference typically relies on analyzing how a model differentiates between member and non-member training data, which often hinges on overfitting. The overfitting effect is considerably reduced without access to the exact graph training data structure, thus attenuating the detection signal.
Similarly, challenges arise in endeavoring to satisfy both \textbf{R2} and \textbf{R3}. The unlearning process (\textbf{R2}) necessitates the identification of specific samples for removal, and evaluating the impact of unlearning on the model is vital to guide this process. However, when the unlearned graph structure is not available (\textbf{R3}), gauging how the removal of samples influences the model becomes formidable, thus impeding the unlearning process. 

\noindent \textbf{Our Contributions.}
To meet requirements \textbf{R1}, \textbf{R3}, and \textbf{R4}, we present a novel proactive misuse detection module that detects graph data misuse without transmitting confidential graph structure data. 
It utilizes an alternative graph without structure during the misuse identification when analyzing the performance difference between misused/benign GNN models. 
To bolster such discernment distributions, we employ a radioactive graph construction method. 
When these graphs are used for training, the distribution difference of the predictions from the misused/benign GNN model is maximized, enhancing the misuse detection performance. 
We introduce a training-graph-free unlearning module to satisfy \textbf{R2}, \textbf{R3}, and \textbf{R4}. 
Our method uses fine-tuning with generated synthetic graphs to carry out unlearning, rather than depending on the exact unlearned graph. 
By using unlearning data samples for fine-tuning, the model increases the loss on these samples, effectively neutralizing their impact.

In summary, we make the following contributions. 
\begin{itemize}
    \item To the best of our knowledge, we present an innovative integrated pipeline framework, called \textbf{GraphGuard}, which is the first practical approach addressing the detection and mitigation of graph data misuse in GNNs within the MLaaS context. 
    \item We define the problem of graph data misuse in MLaaS-deployed GNNs and identify four critical requirements for its mitigation: \textit{detectable}, \textit{remedial}, \textit{data privatization}, and \textit{model agnostic}. 
    \item We introduce an integrated pipeline tailored to \textit{GNNs in MLaaS}, which effectively addresses the outlined requirements. Our approach incorporates a novel misuse detection technique that leverages membership inference augmented with radioactive data, which strengthens the ability to discern between benign and misused model performance. Additionally, we propose an unlearning methodology that 
    employs synthetic graphs to avoid using the confidential graph structure,
    providing an effective means to mitigate the consequences of data misuse.
    \item We conduct extensive experiments on four real-world graph datasets and demonstrate the effectiveness of our design. Specifically, our design, GraphGuard, achieves an almost $100\%$ detection rate across different GNN models, and facilitates unlearning with under $5\%$ accuracy loss. GraphGuard is open-source and available at this repository: \url{https://github.com/GraphGuard/GraphGuard-Proactive}. 
\end{itemize}

\section{Preliminaries}
\label{sec:background}
This section introduces concepts, notations, and the application scenario of our paper, including GNNs, MLaaS, and membership inference in MLaaS. 
\subsection{Graph Neural Networks}

\noindent \textbf{Node Classification via GNNs.} 
GNNs have shown great success in graph analysis tasks. 
This paper considers the node classification task. 
Formally, an attributed graph can be denoted as $G=(A,X)$, where $A\in\{0,1\}^{|{V}| \times |{V}|}$ indicates the graph structure, and $X\in \mathbb{R}^{|{V}|\times d}$ denotes the node features. $V=\{v_1, v_2,..., v_{|{V}|}\}$ represents a set of nodes, and $d$ indicates the dimension of node features (e.g., for node $v_i$, $X_{i} \in \mathbb{R}^{1\times d}$ denotes its features with dimension $d$). Given a set of nodes $V_t \subseteq V$ labeled with $Y_t$, a node classification GNN model $f$ aims to label nodes based on both the graph structure $A$ and node features $X$ in $G$, i.e., $f: V \rightarrow Y$. 

Generally, GNN for node classification has two different learning settings: \textit{transductive} and \textit{inductive}. 
\textbf{(1)} In the transductive setting, the inference graph of the GNN $f$ is the same as the training graph, i.e., all nodes for inference have already been observed at the training time of $f$.
\textbf{(2)} For the inductive setting, the inference graph of $f$ is different from the training graph. 
This setting is similar to traditional learning settings where $f$ learns the knowledge from the training graphs and is used to predict the node label in the unseen graphs.

\noindent \textbf{GNN Model Formulation.}
There are various types of GNN architectures. 
For example, in a graph convolutional network (GCN) model \cite{KipfW17}, each layer aggregates the feature/embedding of both nodes and their neighbors by considering the graph structure. 
Formally, a 2-layer GCN model $f$ is represented as:
\begin{equation}
\label{eqt:GCN}
    f(A,X) = \mathit{Softmax}(\hat{A} \mathit{ReLU} (\hat{A} X W^{(0)})W^{(1)}), 
\end{equation}
where $\hat{A}=\tilde{D}^{-1/2}\tilde{A}\tilde{D}^{-1/2}$ represents the normalized adjacency matrix. 
Given $\tilde{A}=A+I_{|V|}$, $\tilde{D}$ indicates the diagonal degree matrix of $\tilde{A}$ (i.e., $\tilde{D}_{ii}=\sum_j \tilde{A}_{ij}$). 
$W^{(0)}$ and $W^{(1)}$ represent the trainable parameters of $f$.
$\mathit{ReLU}(\cdot)$ and $\mathit{Softmax}(\cdot)$ denote non-linear activation functions. 

\subsection{Machine Learning as a Service}

\noindent \textbf{MLaaS.} MLaaS is trending as an emerging cloud service to train and serve machine learning models~\cite{graphstorm,aws-sagemaker,azure-machine-learning,google-vertex-ai,ibm-wml-zos}. 
It enables model developers to build (\textit{Training Service}) and deploy (\textit{Serving Service}) their models and learning applications to model users conveniently and automatically. 
%
%
A notable framework is Amazon SageMaker~\cite{aws-sagemaker}, which has integrated a popular graph learning tool (i.e., DGL~\cite{dgl_2020}) for the development and deployment of GNNs~\cite{simon_2019}.
%

Model developers, who want to train a GNN model on their graph data, can first schedule (create and configure) a training job through the training service provided by MLaaS platforms.
After that, the training service will receive the training job and launch a training instance that loads the training graph provided by the owners, and then train a GNN $f$ accordingly.
%
Finally, the well-trained $f$ can be downloaded or saved to the cloud storage bucket for later use. 
%
To utilize the serving service, model developers first upload their locally trained GNN $f$ to cloud storage or specify the location of an online GNN $f$, then the Hosting Service can create the API endpoints by utilizing the GNN $f$ and its service configuration. 
These endpoints will remain alive, and model users can then obtain instantaneous predictions by querying these endpoints.

\noindent \textbf{Inductive GNNs in MLaaS.}
Note that, in the inductive setting, the graphs used for the training and inference of GNNs are different. 
Therefore, MLaaS provides APIs for users to query their inference graph during serving. 
Specifically, the model developer uploads only a GNN $f$ to the cloud when creating serving instances. 
GNN users upload their inference graphs and node IDs for predictions during the serving period. 
The serving endpoints then perform prediction on the inference graph and respond to the corresponding labels of the node IDs, respectively.

\subsection{Membership Inference in MLaaS}
\label{sec:MIA_background}

\noindent \textbf{Node-level Membership Inference.}
As a deep learning model, GNN models hunger for large amounts of data to obtain better performance, making the model developer need to gather as much data as possible, raising concerns about data misuse. 
A popular approach to identifying the training data of a given GNN $f$ is membership inference (MI)~\cite{abs-2102-05429,ContiLPX22}. 
In the context of inductive node classification, given a target GNN $f_{\theta^*}$ trained on $G_m=(A_m,X_m)$, whether a node $v_i$ is in the training dataset of $f_{\theta^*}$ (i.e., $\mathtt{membership} = 0$ or $1$) can be 
inferred by
\begin{equation}
\label{eqt:org_mia}
    \begin{aligned}
      \mathtt{membership}=\bigg\{\begin{array}{cc}
                1 & \text{if } \mathtt{mcs}>\eta, \\
                0 & \text{if } \mathtt{mcs} \leq \eta,
            \end{array}
            \mathtt{mcs} = \mathcal{A}(f_{\theta^*}(G)_{v_i}), 
    \end{aligned}
\end{equation}
where $\eta$ indicates a threshold, and $\mathtt{mcs}$ represents the membership confidence score. $f_{\theta^*}(G)_{v_i}$ denotes the prediction vector of node $v_i$ on the queried graph $G$, and $\mathcal{A}(\cdot)$ represents an MI model whose output lies in the range $[0,1]$.

In practical inference, 
since the initiator of MI can not access the training and non-training data of the target GNN $f$, they generally utilize a shadow dataset and shadow model to guide $\mathcal{A}$'s training.
Specifically, to perform membership inference on the target GNN $f_{\theta^*}$ trained on $G_m$, attackers will first gather a shadow dataset $G' = G_m^{'} \cup G_n^{'}$ that shares the same domain as $G_m$. 
Then, they can generate a shadow GNN $f^{'}_{\theta^{'}}$ trained on $G_m^{'}$, followed by obtaining the attack model $\mathcal{A}$ by solving $\max_{\mathcal{A}} d(\mathcal{A}(f^{'}_{\theta^{'}}(G_m^{'})), \mathcal{A}(f^{'}_{\theta^{'}}(G_n^{'})))$, where $\theta^{'} = \arg\min_{\theta} L(f^{'}_{\theta}(G_m^{'}))$ and $d(\cdot,\cdot)$ indicates a distance function. 
As a result, a well-trained attack model $\mathcal{A}$ predicts 1 on member data (i.e., $\mathcal{A}(f^{'}_{\theta^{'}}(G_m^{'}))=1$) while outputs 0 on non-member data (i.e., $\mathcal{A}(f^{'}_{\theta^{'}}(G_n^{'}))=0$). 

\section{Problem Formulation}
In this section, we focus on introducing the application settings and scenarios of our problem and the risk of data misuse in GNNs deployed via MLaaS. 
We first describe the system model by introducing the entities and their roles and goals. 
Then, we present the risk of data misuse in our scenarios. 
And finally, we present the objective for our design to mitigate this misuse problem. 
\subsection{System Model}
\label{sec:system_model}

\begin{table}[t]
    \centering
    \renewcommand\arraystretch{1.3}
    \normalsize
    \tabulinesep=1.1mm
    \caption{Capabilities of Different Entities in Our Application. $\fc$ represents full access, \textcolor{red}{$\hc$} represents access without authorization, while $\ec$ indicates no access. 
    ``Model Developer'' denotes an adversarial model developer who gains access to $A_p$ and $X_p$ without the data owner's authorization. 
    }
    \label{tab:capabilities}
    \begin{tabular}{c|c|c|c|c}
    \toprule
    \multirow{2}{*}{Entity} & \multicolumn{2}{c|}{$G_p$} & \multirow{2}{*}{$G_{m}^{0}$} & \multirow{2}{*}{$f_{\theta^*}$} \\
    \cline{2-3}
     & $A_p$ & $X_p$ & & \\
    \hline
    Data Owner & $\fc$ & $\fc$ & $\ec$ & $\ec$ \\
    MLaaS Server & $\ec$ & $\fc$ & $\ec$ & $\fc$ \\
    Model Developer & \textcolor{red}{$\hc$} & \textcolor{red}{$\hc$} & $\fc$ & $\fc$ \\
    \bottomrule
    \end{tabular}
\end{table}

In this paper, we consider a GNN model developed locally but deployed on an MLaaS server. 
We consider node classification as a basic and popular task for GNNs. 
We focus only on GNNs in inductive settings, which is more common for the model developer to use MLaaS for GNN deployment and serve other GNN users.  
%
Our application scenario includes three distinct entities: 

\noindent \textbf{(1)} \textit{Data Owner:}
Data owners possess graph data, often regarded as their intellectual property. 
They have full access (both knowledge and modification capabilities) to their graph data $G_p = (A_p,X_p)$. 
In our scenario, they only consider the graph structure $A_p$ to be private information, while less sensitive data $X_p$ can be provided to others if necessary (refer to Sec.~\ref{sec:threat_model}). 

\noindent \textbf{(2)} \textit{Model Developer:}
The model developer comprises entities that develop and own a well-trained GNN $f_{\theta^*}$, and deploy $f_{\theta^*}$ on MLaaS platforms, thus severing GNN users by providing them with API to query $f_{\theta^*}$ and charge users and obtain commercial benefit. 
Benign model developer will develop their GNN models (e.g., $f_{\theta^*}$) by gathering and training their model on an authorized training graph $G_{m}^{0}$ (e.g., they buy the graph data from data owners to benefit their GNN training), thus they have full access to both $f_{\theta^*}$ and $G_{m}^{0}$.

\noindent \textbf{(3)} \textit{MLaaS Server:} 
The MLaaS server signifies the entity (e.g.,  Amazon, Microsoft, or Google) providing MLaaS (e.g., cloud computing services) to model developers. 
They help model developers deploy their GNNs (e.g., $f_{\theta^*}$), thus having full access to the model. 
In addition, they may also receive other less sensitive information, such as $X_p$ if necessary. 

We summarize their capabilities in Tab.~\ref{tab:capabilities}. 
In addition, for presentation purposes, we summarize the notation introduced here and in the following sections in Tab.~\ref{tab:notations}.

\subsection{Threat Model}
\label{sec:threat_model}

\mr{
Within the MLaaS ecosystem outlined above, a critical concern is the threat of data misuse during model training, which compromises the IPs of the data owners. 
To scope our discussion, we assume that the data owner is benign, since they have no incentive to undermine their own IPs. 
Similarly, we assume the MLaaS server to be trusty, especially considering that service providers like Amazon and Google have well-established reputations. 
Contrarily, we focus on a scenario in which a model developer is an attacker, acquiring and exploiting unauthorized graph data during their GNN training process, thereby infringing upon the IPs of the data owner. 
}


\noindent 
\mr{
\textbf{Motivations of Data Misuse.}
In this paper, we consider the model developer, who develops a model trained on unauthorized data and deploys it via MLaaS, as an attacker. 
This model is trained using meticulously gathered training data without the data owner's consent to exhibit superior performance. 
Such proficiency allows the model developer to gain commercial benefits by selling predictions through MLaaS, while avoiding costs associated with procuring the training data. 
This kind of unauthorized data usage severely undermines the IP rights of the data owner. 
It is worth noting that such a data misuse problem is not rare in practice. 
OpenAI, for example, faced lawsuits from various authors who claimed that their data were integrated into ML models without permission~\cite{Frenkel_Thompson_2023}. 
Similarly, artists have raised concerns about LensaAI, a tool that is alleged to train models using their artwork to reproduce specific styles, contending that it infringes on their rights~\cite{Mattei_2022}. 
}

\begin{table}[]
    \small
    \centering
    {
        \caption{Notations and Explanations.}
        \label{tab:notations}
        \begin{tabular}{c p{6.5cm}}
        \toprule
          Notation  & Explanation \\
        \toprule
        $G_p$ & Graph owned by data owner, $G_p= (A_p,X_p)$ \\
        $\hat{G}_p$  & 
        Queried graph, $\hat{G}_p=(\hat{A}_p,X_p)$ and $\hat{A}_p = I$
        \\
        $G_m$ & Training graph of the model developer \\
        $G_m^0$ & Authorized training graph ($G_p \cap {G}_m^{0} = \emptyset$) \\
        $G_m^1$ & 
        Unauthorized training graph ($G_p \cap G_m^{1} \neq \emptyset$)
        \\
        $\tilde{G}_p$ & Synthetic unlearning graph \\
        $\tilde{G}_r$ & Synthetic remaining graph \\
        \midrule
        $A_p$ & Adjacency matrix of $G_p$ \\
        $\hat{A}_p$ & Adjacency matrix of $\hat{G}_p$ ($\hat{A}_p = I$) \\
        $X_p$ & Node attributes of $G_p$ and $\hat{G}_p$\\
        $X_{m}^{0}$ & Node attributes of $G_m^0$\\
        $Y_p$ & Labels of the nodes in $G_p$ and $\hat{G}_p$ \\
        \midrule
        $f_{\theta^*}$ & GNN trained by model developer\\
        $f_{\theta^*_0}$ & GNN trained on authorized graph $G_m^0$ \\
        $f_{\theta^*_1}$ & GNN trained on unauthorized graph $G_m^1$ \\
        $f_{\widetilde{\theta^{*}}}$ & Unlearned GNN \\
        $f_{\theta^{p}}$ & Pre-trained surrogate model \\
        $g$ & Graph generation model \\
        \midrule
        $L$ & Loss function during GNN training\\
        $\mathcal{A}$ & Membership inference attack model \\
        %
        \bottomrule
        \end{tabular}
    }
\end{table}

\noindent 
\mr{
\textbf{Attacker Capabilities in Data Misuse.}
Considering the data misuse attack mentioned earlier, we attribute two capabilities to the attacker:
1) Obtaining data from the data owner without authorization. This unauthorized access often stems from lapses in data management or oversight by the data owner. For example, several medical companies have experienced data breaches due to inadequate data management or system misconfigurations~\cite{Burky_2022,Gatlan_2023,Geer_2019}.
2) Deploy a model within MLaaS while concealing its illicit use of unauthorized data. Since the training process is executed locally by the model developer, the MLaaS server lacks visibility into their local training activities. 
This makes it challenging to determine whether a model was trained on unauthorized data. 
}

\mr{
Based on the above attack setting, we formally provide the definition of training graph data misuse for GNNs in MLaaS as follows:
}
\begin{definition}
    \textit{[Data Misuse for GNNs in MLaaS]}
    Consider a model developer who intends to develop a well-trained GNN model using an authorized graph $G_{m}^{0}$. 
    In the case of \textit{data misuse}, the model developer, in addition to $G_{m}^{0}$, intentionally or occasionally collects an unauthorized graph $G_p$, and uses $G_m^{1}= G_{m}^{0} \cup G_p$ as a training set to construct the GNN $f_{\theta^*}$ in a local setting. 
    Subsequently, the model developer provides the trained model $f_{\theta^*}$ to the MLaaS server for deployment and the GNN model service, without disclosing their use of unauthorized data. 
\end{definition}


\noindent
\mr{
\textbf{Remarks. }
\textit{1) Node Classification.}  
In this paper, we focus primarily on node classification as it is a foundational task in GNN applications. Given that graph classification and link prediction also rely on node embeddings, our methods can be readily extended to these tasks. 
A detailed discussion on how our design can be broadened to include other GNN tasks can be found in Sec.~\ref{sec:discussion}.
}

\noindent 
\mr{
\textit{2) Inductive Settings.}  
We focus on attacks targeting inductive GNNs, where the training and inference graphs are distinct. 
This setting makes detecting graph misuse challenging due to the inaccessibility of the training process in the MLaaS setting. 
In transductive settings, the inference graph is the same as the training graph, allowing the benign server an immediate opportunity to determine if the inference graph is unauthorized and has been used for training without consent.
}


\subsection{Design Requirements}
\label{sec:design_req}
This paper aims to develop a framework to identify and mitigate data misuse in the context of GNN in MLaaS. 
Our method should be suitable for the MLaaS scenario that meticulously addresses the settings of each participating entity. 
Specifically, our design requirements are as follows.
\begin{itemize}
    \item \textbf{R1}--\textbf{Detectable:} Our method must facilitate the detection of data misuse, allowing the identification of unauthorized GNN models deployed in MLaaS that have been trained by misusing graph data without authorization, even when the training process of the data misuse is less transparent (i.e., locally performed by the malicious model developer). 
    \item \textbf{R2}--\textbf{Remedial:} Our framework should enable the MLaaS server to initiate or coordinate the unlearning process in cases where data misuse is detected. This involves either performing necessary actions to remove the impact of unauthorized data usage or protecting the data owner's forgotten right. 
%
    \item \textbf{R3}--\textbf{Data Privatization:} A critical aspect of our design in MLaaS settings is protecting the data privatization of every entity throughout our mitigation process. Our system has multiple entities, and they will have their own data that will not be shared with others. For example, the data owner's graph structure should not be exposed to other entities during the mitigation process due to the data privatization concern or data usage agreements. This means that the data owner is not expected to share their graph structure with the MLaaS server or the model developer. Meanwhile, the MLaaS server and model developer will not share the GNN model with the data owner, and model developer will not share their training graph data with the server (see their knowledge in Tab.~\ref{tab:capabilities}).
    \item \textbf{R4}--\textbf{Model Agnostic:} We do not make any assumptions or requirements for the three entities. Specifically, for the model developer, we do not assume any additional process during the training of the GNN model. For MLaaS, we do not require specified architectures for the deployment of GNN models. Namely, our design can be applied to GNNs with a general model architecture in the context of MLaaS. 
\end{itemize}




\noindent \textbf{Remark for Data Privatization.} 
Here we specify the privatized data for each entity that we considered in this paper. 
For the model developer and MLaaS server, well-trained GNN should be considered privatized data and not provided to the data owner or general GNN user. 
In addition, the graph structure for the data owner's and the model developer's training graphs should also be considered privatized data and not be provided to any other entities. 
This is a common scenario in GNN systems and has been the subject of previous studies on GNN security~\cite{he2021stealing,Choquette-ChooT21,wu2022linkteller,0001MMBS22,abs-2205-07424}.
Collecting training graphs and building a GNN model often requires a large amount of human, computing, and economic resources, and thus a model developer is concerned about the privatization of its training graphs and GNNs and would not provide them to others. 
Meanwhile, the data owner should keep the graph structure information privatized, and never be provided to other entities in our system. 

\mr{
Note that our design focuses only on the graph structure. 
Given that graph data primarily represent connections and interactions between node entities, the graph structure tends to contain more sensitive information in GNN applications~\cite{wu2022linkteller}. 
A practical example could be a company managing a private transaction graph, where nodes represent individuals and node attributes (i.e., features) represent publicly available basic profiles. 
Edges represent sensitive transactions among individuals. 
These edges are considered the property of the company, and they may express concern over the privatization of their graph structure data. 
Another example could be a GNN used for PPI in drug discovery, where nodes represent proteins and node attributes represent well-known properties~\cite{LiuYCJ20,wang2023assessment}. 
The edges represent interactions among proteins, which should be obtained by extensive experimentation~\cite{melo2016machine,mrowka2001there}.
We also discuss how our design can be extended to also protect the node attributes in Sec.~\ref{sec:discussion}.
}

\section{GraphGuard}

\begin{figure*}
    \centering
    \includegraphics[width=\textwidth]{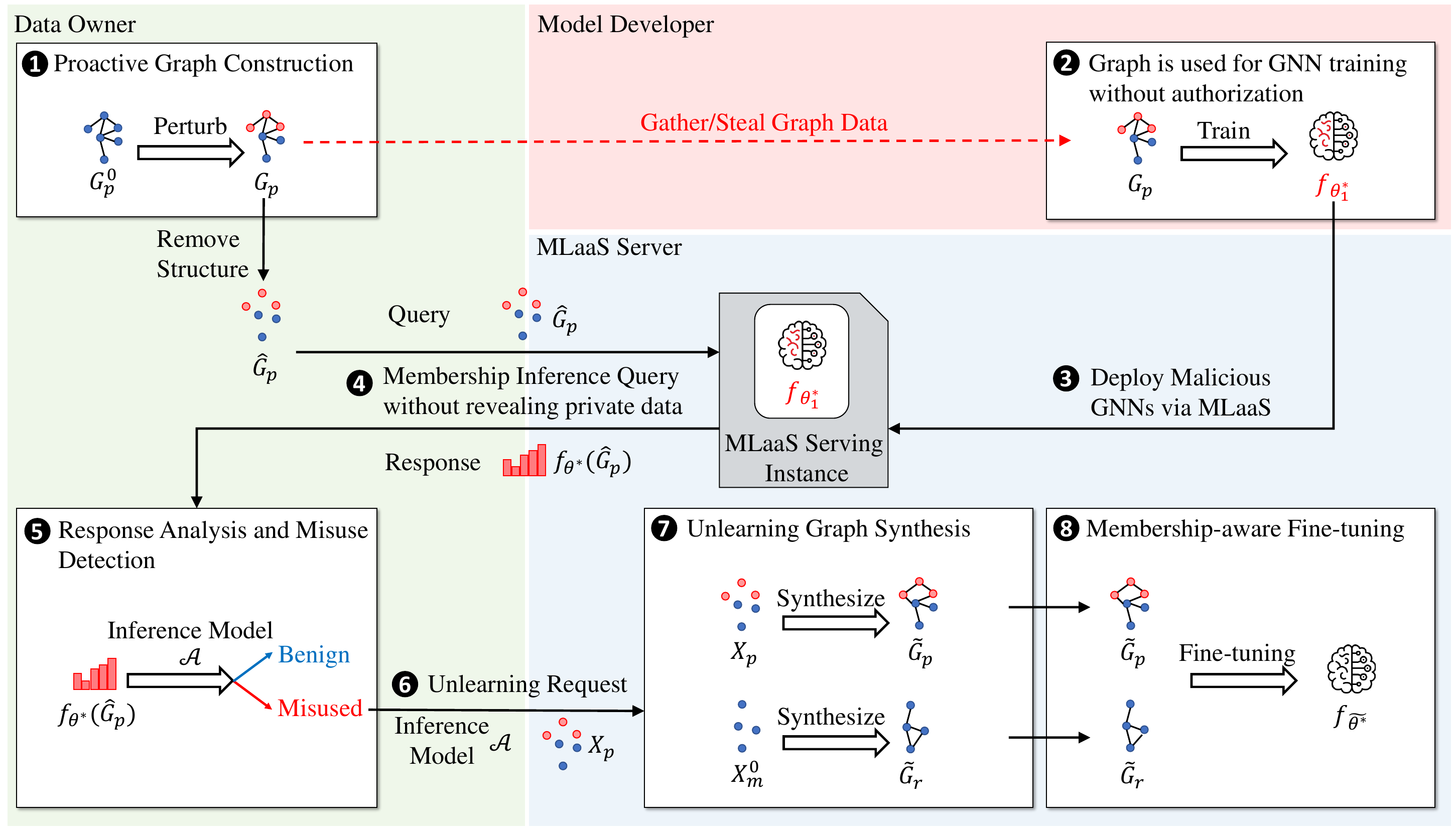}
    \caption{\mr{Overall Pipeline of Our Graph Data Misuse Mitigation Framework.} 
    }
    \label{fig:design_overall}
\end{figure*}

\mr{
To deal with the training graph misuse problem in GNNs, in this paper, we introduce an integrated framework, GraphGuard, which can both detect and mitigate the data misuse and is suitable in the context of MLaaS. 
Specifically, we split the data misuse mitigation into two stages of GraphGuard (i.e., detecting data misuse and reducing its effects) to match the requirements R1 and R2, followed by proposing two respective components, misuse detection and unlearning. 
Moreover, we make relevant adjustments or enhancements to the misuse detection and unlearning components to address other requirements (i.e., R3 and R4).
%
%
A detailed flow chart of this process is presented in Fig.~\ref{fig:design_overall}.
In this section, we will introduce each of these two components by first defining their overall goals, technical challenges when considering the practical MLaaS scenarios, design intuition, and then the detailed methodology. 
}

\subsection{Graph Data Misuse Detection}
\label{sec:proactiveMIA}
Graph data misuse detection concentrates on determining whether misuse has occurred in the context of a suspect GNN model deployed on an MLaaS platform. 
Within our framework, data owners will perform this detection task. 
Since they do not have direct access to the suspect GNN model, they may issue queries to the MLaaS server and ask them to perform inference on these queries and generate a response to assist in their misuse detection. 
Formally, the objective of our detection module can be elucidated as follows: 
\begin{definition}
\textit{[Graph Data Misuse Detection for GNNs in MLaaS]}
Assuming a suspect GNN $f_{\theta^*}$, which is trained on a graph $G_m$, is now deployed in MLaaS. With only the ability to query $f_{\theta^*}$, the data owner aims to identify if graph data $G_p=(A_p, X_p)$ was used in training $f_{\theta^*}$ (i.e., whether $G_m \cap G_p = \emptyset$) without divulging private information $A_p$.
\end{definition}

\mr{
%
%
It can be found that the above detection has a similar objective as the membership inference attack we introduced in Sec.~\ref{sec:background}, but with additional constraints on preventing the disclosure of information regarding $A_p$. 
Therefore, we can consider our detection as a membership inference problem, while adding noise to the membership inference query samples to protect sensitive information in $A_p$. 
Namely, to prevent the private sample from being disclosed, rather than querying precise samples $G_p=(A_p, X_p)$, our membership inference process can query a perturbed version of $G_p$ to the target model deployed on the server. 
}

\noindent
\mr{
\textbf{Challenges and Design Intuitions.} 
However, introducing noise to the query samples makes membership inference considerably less effective. 
Note that membership inferences are based on the analysis of the responses to the query sample~\cite{abs-2102-05429}.
Specifically, whether a specific sample has been used for training is determined based on whether the target model overfits the query samples~\cite{0001MMBS22,CarliniCN0TT22}. 
Therefore, when the query samples are perturbed, even if they are training members, the target model will exhibit reduced overfitting on them, making them harder to accurately classify as members. 
}

\mr{
To address this challenge, we draw inspiration from the concept of radioactive data, as first introduced in~\cite{SablayrollesDSJ20}. Instead of relying solely on overfitting, we advocate for constructing a \textit{radioactive graph} to facilitate the detection of graph misuse. 
Specifically, we introduce optimized perturbations to the original graph data, yielding radioactive graphs. 
We expect that any model trained using our radioactive graph will bear a unique mark. 
This mark can render the output distributions of a vanilla model and a data-misused model trained on the radioactive graph distinctly different.
}



\noindent
\mr{
\textbf{Design Overview.}
By utilizing the radioactive graph for membership inference, our data misuse detection process will comprise two phases, including an initialization phase for the radioactive graph construction, and a detection phase for membership inference, which can be succinctly outlined as follows:
}

\noindent
\mr{
\textit{Initialization Phase:}
\begin{itemize}
    \item [1.] The data owner converts their original data into a radioactive graph locally via our radioactive graph construction algorithm.
\end{itemize}
}

\mr{
After the radioactive graph is generated and stored, the potential misuse occurs as how we discussed in Sec.~\ref{sec:threat_model}. 
Specifically, the radioactive graph may be illicitly acquired and utilized by adversarial model developers to augment their models without the data owner's consent. 
These developers then deploy their model to MLaaS platforms and market their prediction services.
}

\mr{
When the data owner believes that a model on MLaaS might be exploiting their data for training (e.g., due to its task relevance), they initiate the detection phase.
}

\noindent
\mr{
\textit{Detection Phase:}
\begin{itemize}
    \item [2.] The data owner crafts a query about their radioactive graph, while ensuring the omission of structural details (i.e., removing all edges, querying only individual nodes coupled with their attributes).
    \item [3.] The MLaaS performs inference on the suspected GNN model with the query graph and subsequently transmits the results to the data owner. 
    \item [4.] The data owner evaluates the received response to discern if the operational model on MLaaS was trained using the radioactive graph data without the data user's authorization.
\end{itemize}
}

\noindent 
\mr{
\textbf{Radioactive Graph Construction.}
With the aforementioned workflow in place, our goal is to construct the radioactive graph. 
We first define the optimization problem of radioactive graph construction and then present how to solve it. 
}

\mr{
Our radioactive graph is designed to render the output distributions of the vanilla model and data misused model difference. 
A common strategy for assessing such differences is to follow similar strategies as membership inference and calculate the prediction disparity $D$ of these outputs. 
Specifically, as shown in Sec.~\ref{sec:MIA_background}, the distribution difference for member and non-member data can be calculated as: $d(\mathcal{A}(f_{\theta^{*}}(G_m)), \mathcal{A}(f_{\theta^{*}}(G_n)))$.
Accordingly, in the context of increasing the difference by the radioactive graph, the objective of the output distribution can be represented as: 
\begin{equation}
\label{eq:rgraph_op}
\begin{aligned}
\max_{G_p}\ & d(\mathcal{A}(f_{\theta^{*}_1}(\hat{G}_p)), \mathcal{A}(f_{\theta^{*}_0}(\hat{G}_p))), \\
s.t. \ & \theta^{*}_{1} =\arg \min_{\theta} L(f_{\theta}(G_m^{1})),\\
& \theta^{*}_{0} =\arg \min_{\theta} L(f_{\theta}({G}_m^{0})),\\
\end{aligned} 
\end{equation}
where $\mathcal{A}(\cdot)$ is a membership inference attack model,  $d(\cdot,\cdot)$ indicates a distance function, $\hat{G}_p = (\hat{A}_p,X_p) = (I_{|V|},X_p)$ is the queried graph without graph structure information (i.e., all nodes in $\hat{G}_p$ are isolated),  $G_m^{1}$ and $G_m^{0}$ represent two different versions of $G_m$, namely, $G_m^{1}$ indicates the $G_m$ that (partially) includes $G_p$ (i.e., $G_p \cap G_m^{1} \neq \emptyset$) while $G_m^{0}$ denotes the $G_m$ that excludes $G_p$ (i.e., $G_p \cap {G}_m^{0} = \emptyset$). 
}
\mr{
Note that, obtaining $\hat{G}_p$ from (\ref{eq:rgraph_op}) is non-trivial and (\ref{eq:rgraph_op}) contains unspecified functions such as $\mathcal{A}(\cdot)$ and $d(\cdot,\cdot)$. 
Therefore, we convert the optimization problem in (\ref{eq:rgraph_op}) as follows:
\begin{equation}
        \min_{X_p}\ ||f_{\theta^{p}}^{e}(G_p)-f_{\theta^{p}}^{e}(\hat{G}_p)||_2 - L(f_{\theta^{p}}(\hat{G}_p),Y_p),
        \label{eqt:PMIA_converted_opt}
\end{equation}
where $L$ is GNN's training loss function, $f_{\theta^{p}} = f_{\theta^{p}}^{c} \circ f_{\theta^{p}}^{e}$ is a pre-trained surrogate model (e.g., any publicly available GNN model performing similar tasks, which is consistent with prior works' assumption on graph unlearning~\cite{Chen000H022}). 
$f_{\theta^{p}}^{e}$ and $f_{\theta^{p}}^{c}$ indicate the classifier and encoder of $f_{\theta^{p}}$, respectively. $\circ$ represents the composition of functions.
%
This optimization can be efficiently solved using the gradient descent method. 
We provide the proof of the above conversion in App.~\ref{app:radioactive_construction_proof}. 
A step-by-step procedure for constructing the radioactive graph is delineated in Algorithm~\ref{alg:radioactive_graph_construction}. 
}



\begin{algorithm}[t]
\caption{Radioactive Graph Construction}\label{alg:radioactive_graph_construction}
\begin{flushleft}
\hspace*{\algorithmicindent}\textbf{Input:} \\
\hspace*{\algorithmicindent} $G_{p}^{0} = (A_p, X_{p}^{0})$ Initial graph owned by the data owner, \\ 
\hspace*{\algorithmicindent} $Y_p$ Labels of the graph owned by the data owner, \\
\hspace*{\algorithmicindent} $N$ Maximum construction epoch, \\
\hspace*{\algorithmicindent} $k$ Feature construction step, \\
\hspace*{\algorithmicindent} $f_{\theta^{p}} = f_{\theta^{p}}^{c} \circ f_{\theta^{p}}^{e}$ A pretrained GNNs. \\
\hspace*{\algorithmicindent}\textbf{Output:} \\
\hspace*{\algorithmicindent} $G_{p} = (A_p, X_{p})$ Radioactive graph. \\
\end{flushleft}
\begin{algorithmic}[1]
\For {$n \text{ \textbf{from} } 0 \text{ \textbf{to} } N-1$}
    \State $G_{p}^{n+1} = (A_p, X_{p}^{n+1}) \leftarrow G_{p}^{n} = (A_p, X_{p}^{n})$ 
    \State $L_{opt}$ $\leftarrow$ $||f_{\theta^{p}}^{e}(G_{p}^{n})-f_{\theta^{p}}^{e}(\hat{G}_{p}^{n})||_2 - L(f_{\theta^{p}}(\hat{G}_{p}^{n}),Y_p)$
    \State $\Delta_{L}$ $\leftarrow$ $\frac{\partial L_{opt}}{\partial X_{p}^{0}}$
    \State $X_{p}^{'}$ $\leftarrow$ Selecting the top-$k$ largest $X_{p}[i]$ w.r.t. $|\Delta_{L}[i]|$
    \For {$X_{p}[i] \in X_{p}^{'}$}
        \If {$\Delta_{L}[i] > 0$}
            \State $X_{p}^{n+1}[i] \leftarrow 0$
        \Else
            \State $X_{p}^{n+1}[i] \leftarrow 1$
        \EndIf
    \EndFor
\EndFor
\end{algorithmic}
\end{algorithm}

\noindent \textbf{Discussion.}
\textit{a) Privacy Risk of Querying Node Attribute.}
A potential concern relates to link stealing attacks, which can leverage node prediction results to deduce the graph structure. Such querying and responding could potentially introduce additional vulnerability to link leakage~\cite{he2021stealing,wu2022linkteller}. However, it should be noted that all existing link-stealing attacks~\cite{he2021stealing,wu2022linkteller} require inference results on a graph that includes a private link. On the contrary, our approach involves inference on a graph without a private link. Additionally, we introduce perturbations into the node attributes, thereby reducing the effectiveness of link recovery methods that use graph structure learning. 

\noindent \textit{b) Distinction from Backdoor and Poisoning Approaches.}
Other works have proposed the injection of data into the training dataset~\cite{abs-2006-11890,ZhangJWG21,ZhangCS0X22}. However, our approach has distinct settings and objectives. Existing approaches assume that an attacker can influence the training process, while the data owner remains unaware if their data have been used during training. Our goal, on the other hand, is to amplify the difference between member and non-member data, while other approaches aim to cause misclassification or reduce model performance.

\noindent \textit{c) Distinction from Radioactive Data.} 
Other work~\cite{SablayrollesDSJ20} also proposed to perturb data for detection.
However, they focus only on linear models and Euclidean data. 
Our design considers GNNs which are nonlinear models and used for analyzing graph structure data. 
Their design is not applicable to our case since 1) their detection heavily relies on the assumption that the model is linear, while common GNN has nonlinear activation layers; 2) their design focus on attribute and ignore the graph structure information.

\noindent
\mr{
\textit{d) Distinction from Model Watermarking.} 
Other studies have suggested embedding watermarks into the ML model during its training phase. 
Note that these techniques serve distinct objectives compared to our design: they aim to protect the IP of the model developer, not that of the data owner. 
As a result, their settings differ from ours. 
Specifically, watermarking is implemented by the model developer, who often has full access to the training period. 
In contrast, in our context, the data misuse detection is carried out by the data owner, who does not have access to the training process. 
}

\subsection{Graph Data Misuse Mitigation}
\label{sec:unlearing}

\mr{
This component focuses on the implementation of unlearning in the infringing GNN model to eliminate the influence of unauthorized training data samples. 
The MLaaS server will undertake this task and perform unlearning after receiving a request from the data owner regarding a deployed infringing GNN model. 
Since the MLaaS servers can access and modify the GNN model via retraining or fine-tuning, but do not have access to the training graph data, they may ask the data owner, who issued the unlearning request, to provide less sensitive information (i.e., node attributes) to facilitate the unlearning process. 
Specifically, we formalize the objective of our second component, i.e., graph unlearning, as follows:
}
\begin{definition}
    \textit{[Training-graph-free Unlearning for GNNs in MLaaS]} 
    Given that $G_m^{0}$ indicates the authorized graph, a GNN $f_{\theta^*}$ is identified as data-misused GNN $f_{\theta^{*}_{1}}$ when it is trained in $G_m^{1}$, where $G_m^{1}= G_m^{0} \cup G_{p}$, and $G_{p}=(X_{p},A_{p})$ represents an unauthorized graph. 
    Without accessing the sensitive graph structure of $G_{p}$ and the authorized graph $G_{m}^{0}$ (i.e., $A_{p}$ and $A_{m}^{0}$), the MLaaS server aims to reconstruct an unlearned GNN $f_{\widetilde{\theta^{*}}}$, approximating the benign GNN $f_{\theta^{*}_{0}}$ trained in $G_m^{0}$. 
\end{definition}

\noindent 
\mr{
\textbf{Challenges and Design Intuitions.} 
In the above mitigation process, a key stipulation is that, the graph structure, as the private information, must remain undisclosed to the server. 
This presents significant challenges compared to previous unlearning studies. 
In particular, most existing processes~\cite{Chen000H022,chien2022certified,cheng2023gnndelete,WuYQS0023} require that the entities performing the unlearning can access the unlearning samples so that these samples guide which information to forget. 
In the absence of these samples, the server remains clueless about which specific knowledge should be expunged from the GNN model undergoing unlearning, and making ordinary unlearning methods inapplicable. 
}

\mr{
To deal with this challenge, without dispatching the unlearning samples directly to the server, we propose executing the unlearning with fine-tuning based on synthetic graphs self-generated by the server. 
The graph synthesis is based on statistical knowledge derived from a membership inference attack model, augmented by some auxiliary knowledge (for instance, less sensitive node attributes supplied by the data owner). 
This approach eschews the direct sharing of sensitive graph structure data. 
Nonetheless, the unlearning remains directed by the information embodied in the membership inference model. 
Note that the membership inference model encapsulates the knowledge the target model has acquired, which precisely contains what we intend the target model to forget. 
}

\begin{figure}
    \centering
    \includegraphics[width=0.48\textwidth]{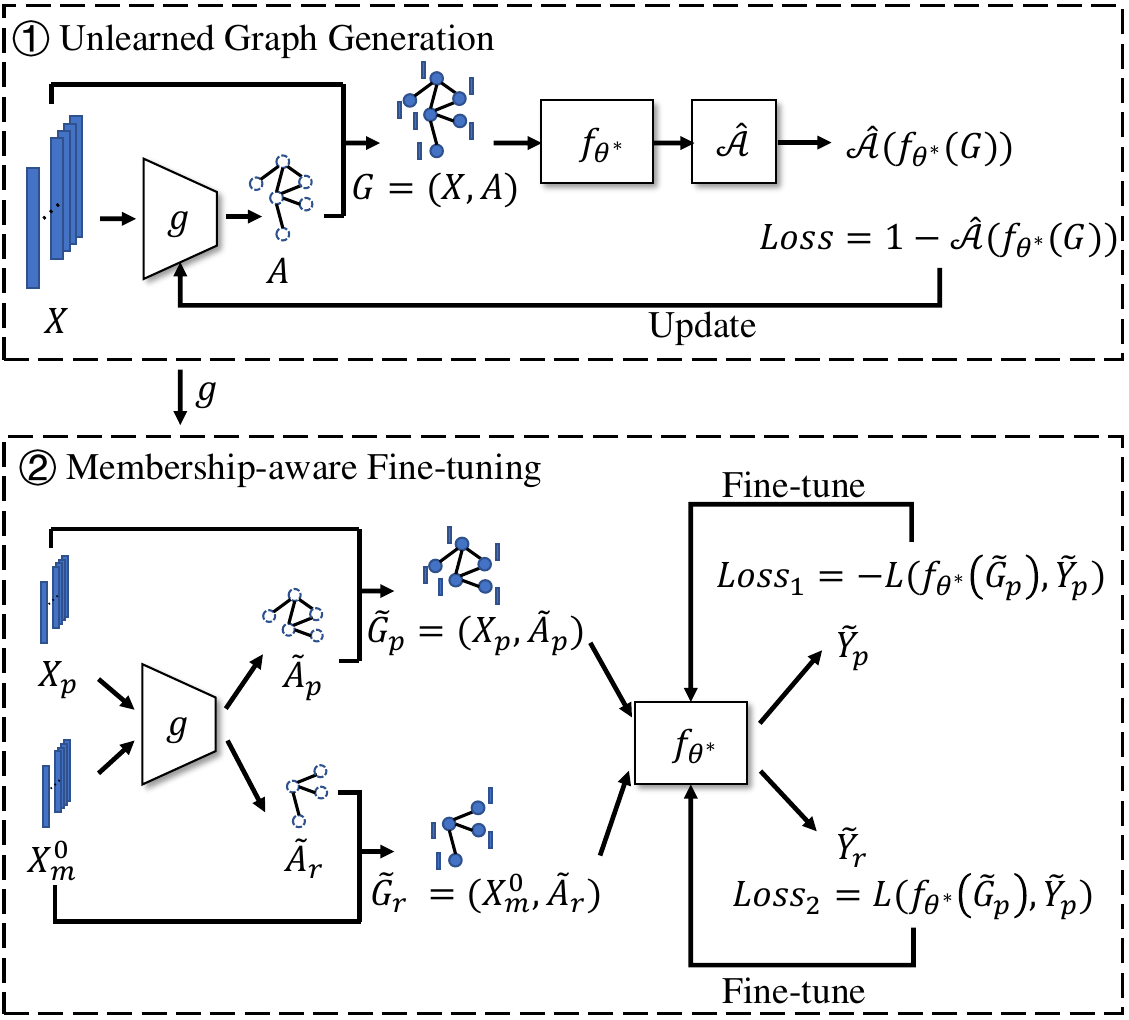}
    \caption{Overall Architecture of Training-graph-free Unlearning.}
    \label{fig:unlearning}
    \vspace{-5pt}
\end{figure}

\noindent 
\mr{
\textbf{Detailed Methodologies.}
Building upon the above intuitions, our mitigation strategy unfolds in two distinct stages: the generation of the unlearned graph and membership-aware fine-tuning.
}
%
%
Specifically, our unlearning module consists of two steps: unlearned graph generation and membership-aware fine-tuning. 
The overall architecture of our proposed solution is illustrated in Fig.~\ref{fig:unlearning}. 

\noindent \textit{a) Unlearned Graph Generation.}
\label{sec:synthesis_graph_alg}
Given access to less-sensitive node attributes, the MLaaS server will request the data owner to upload the $X_p$ (i.e., node attributes of the nodes to be unlearned) upon receiving an unlearning request. 
With these attributes at its disposal, which can be viewed as isolated nodes, the server will construct their interconnections. 
An intuitive design would be to directly connect nodes with similar node attributes due to the homophily of graph data. 
However, this approach falls short as it only considers the knowledge of the node attribute, 
while ignoring the original connection between these nodes.
Consequently, performing unlearning on these generated graphs
may lead to inadequate unlearning. 

To this end, we propose to utilize the knowledge from the target GNN $f_{\theta^{*}}$. Concretely, we train a model $g = g^{c} \circ g^{e}: X \rightarrow (A,X)$ to connect nodes and then generate a graph $G=(A,X)$, where $g^{c}$ and $g^{e}$ indicate the classifier and encoder of $g$, respectively. Given access to the node features $X$, the inference model $\mathcal{A}$ and the target GNN $f_{\theta^{*}}$, $g$ is trained by the following optimization.
\begin{equation}
    \min_{g} 1 - \mathcal{A}(f_{\theta^{*}}(g(X))), 
    \label{eqt:unlearning_encoder}
\end{equation}
where an edge exists between two nodes $v_i$ and $v_j$ (i.e., $A_{i,j} = 1$) when $||g^{e}(v_i) - g^{e}(v_j)||_{2}\leq \epsilon_u$, $\epsilon_u$ indicates a predefined threshold.
The intuition of Equation (\ref{eqt:unlearning_encoder}) is that, the inference model $\mathcal{A}$ outputs $1$ if the generated graph $g(X)$ is overfitted by the target unlearning model $f_{\theta^{*}}$.
Note that, in addition to the node features $X_p$ of the unlearning nodes, we also request the model developer to provide $X_{m}^{0}$ (i.e., node features of the authorized training graph) and involve it in the training of $g$, which facilitates the following fine-tuning to maintain the performance of $f_{\theta^{*}}$. 
With the well-trained $g$, we obtain the unlearning graph $\tilde{G}_p=g(X_p)$ and the remaining graph $\tilde{G}_r=g(X_r)$ for the following membership-aware fine-tuning.

\noindent \textit{b) Membership-aware Fine-tuning.} 
\label{sec:membership_aware_unlearning_alg}
The objective of unlearning is to make the target model $f_{\theta^{*}}$ forget the unlearned graph samples while maintaining performance on the remaining graph samples. 
In this paper, we proposed using the fine-tuning methodology to conduct GNN unlearning, which is motivated by the fact that a GNN model learns by updating its weights to minimize training loss in its training graph. 
Specifically, to maintain performance when unlearning, the unlearned GNN $f_{\widetilde{\theta^{*}}}$ is obtained by
\begin{equation}
\widetilde{\theta^{*}} = \arg \min_{\theta} = L(f_{\theta}(\tilde{G}_r)) - \alpha L(f_{\theta}(\tilde{G}_p)),
\end{equation}
where $L$ represents the training loss function, $\alpha$ indicates a pre-defined hyper-parameter to balance the unlearning request and GNN utility. 
Training loss is calculated by comparing the prediction results with the initial prediction labels generated by $f_{\theta^{*}}$ before fine-tuning. 
Note that $\alpha$ can be adjusted based on the size ratio between $\tilde{G}_r$ and $\tilde{G}_p$. For example, a smaller unlearning graph $\tilde{G}_p$ might require lower values of $\alpha$. 
In our design, we use $0.5$ to balance the utility of the model and the effectiveness of unlearning. 


\noindent 
\mr{
\textbf{Remarks.} 
The generation of our unlearned graph does not necessitate the exact reconstruction of private edges. 
Instead, both the unlearning and remaining graph data are synthesized exclusively from node attributes, removing edge information (see Fig.~\ref{fig:unlearning} and Sec.~\ref{sec:synthesis_graph_alg}). 
And our unlearned graph generation model also does not rely on the private graph structure, but utilizes latent patterns deriving from the inference attack model and node attributes (see Equation~(\ref{eqt:unlearning_encoder}) and Sec.~\ref{sec:synthesis_graph_alg}). 
}

\subsection{Combination}
\label{sec:design_overall}


As shown in Fig. \ref{fig:design_overall}, we propose our graph misuse mitigation framework by combining the \textit{proactive misuse detection} (i.e., Sec.~\ref{sec:proactiveMIA}) and \textit{training-graph-free unlearning} (i.e., Sec.~\ref{sec:unlearing}) together. 
Specifically, our framework mainly includes two phases, as shown in the following.


\noindent \textbf{1)} The data owner will first construct the proactive graph $G_p$ by perturbing the node attributes of the benign graph $G_p^0$ by our radioactive graph construction algorithm (described in Sec.~\ref{sec:proactiveMIA}).

\noindent \textbf{2)} 
Data misuse occurs, where model developer may misuse $G_p$ intentionally or occasionally where $f_{\theta^*}=f_{\theta^{*}_{1}}$ is trained on $G_{m}^{1} = G_p \cup G_{m}^0$ (as described in Sec.~\ref{sec:threat_model}).

\noindent \textbf{3)} The model developer deploy the data-misused model $f_{\theta^{*}_{1}}$ via MLaaS server.  


\noindent \textbf{4)} 
When data owners' concerns on data misuse of a suspect GNN $f_{\theta^{*}_{1}}$ arise, they will submit membership inference queries on $\hat{G}_p$ (whose structure has been removed for privacy consideration) to the MLaaS server. 
The MLaaS server processes the inference employing the targeted GNN $f_{\theta^{*}_{1}}$, subsequently relaying the prediction results from $f_{\theta^{*}_{1}}(\hat{G}_p)$ to the data owner. 
%


\noindent \textbf{5)} Data owners, based on the confidence scores of these query results (i.e., $f_{\theta^{*}_{1}}(\hat{G}_p)$), determine whether data misuse occurred. 

\noindent \textbf{6)} Once identifying data misuse 
The data owner will request the MLaaS server to perform unlearning by providing their node attributes $X_p$ and inference model $\mathcal{A}$ for their unlearned graph. 

\noindent \textbf{7)} The MLaaS server will also request the GNN model developer to provide the node attributes from their authorized training graph $X_{m}^{0}$. 
And the MLaaS server generates the unlearned graph $\tilde{G}_p$ based on $X_p$, and the remaining graph $\tilde{G}_r$ based on $X_{m}^{0}$ (as our method in Sec.~\ref{sec:synthesis_graph_alg}). 

\noindent \textbf{8)} The MLaaS server can finally obtain an unlearned GNN $f_{\widetilde{\theta^{*}}}$ based on both $\tilde{G}_p$ and $\tilde{G}_r$ by following our unlearning method (as described in Sec.~\ref{sec:membership_aware_unlearning_alg}).

Among the above, Steps 1, 4, and 5 are our \textit{proactive misuse detection} module;
Steps 2 and 3 are the data misuse threat;
and Steps 6 to 8 are our \textit{training-graph-free unlearning} module. 

\section{Experiments and Evaluations}
\label{sec:exp}

\begin{table}[t]
\normalsize
\tabulinesep=1.1mm
    \centering
    \caption{GNN Training Graph Statistics.}
    \label{tab:data_statistic}
    \begin{tabular}{c|rrr}
    \toprule
        Datasets &  \# Nodes   & \# Edges &  \# Attributes   \\
    \toprule
      Cora & 2,708 & 5,429 & 1,433 \\
      Citeseer & 3,327 & 4,732 & 3,703 \\
      Pubmed & 19,717 & 44,338 & 500 \\
      Flickr & 89,250 & 899,756 & 500 \\
    \bottomrule
    \end{tabular}
\end{table}

\subsection{Experiment Setup}

\begin{table*}[t!]
\centering
\normalsize
\tabulinesep=1.1mm
\caption{Comparison of AUC between Our Method and Baseline Membership Inference.}
\label{tab:comparison_detection_baseline}
\begin{threeparttable}
\begin{tabularx}{\textwidth}{c|*{12}{>{\centering\arraybackslash}X}}
\hline
\multirow{2}{*}{} & \multicolumn{3}{c}{GCN} & \multicolumn{3}{c}{GraphSage} & \multicolumn{3}{c}{GAT} & \multicolumn{3}{c}{GIN} \\ \cline{2-13}
 & Baseline & Ours & $\Delta$ & Baseline & Ours & $\Delta$ & Baseline & Ours & $\Delta$ & Baseline & Ours & $\Delta$ \\ \hline
Cora & 0.874 & 0.999 & \textbf{$\uparrow$0.125} & 0.864 & 0.999 & \textbf{$\uparrow$0.135} & 0.927 & 1.0 & \textbf{$\uparrow$0.073} & 0.857 & 1.0 & \textbf{$\uparrow$0.143} \\

Citeseer & 0.711 & 0.999 & \textbf{$\uparrow$0.288} & 0.822 & 1.0 & \textbf{$\uparrow$0.178} & 0.723 & 0.999 & \textbf{$\uparrow$0.276} & 0.767 & 1.0 & \textbf{$\uparrow$0.233} \\

Pubmed & 0.906 & 1.0 & \textbf{$\uparrow$0.094} & 0.902 & 1.0 & \textbf{$\uparrow$0.098} & 1.0 & 1.0 & \textbf{0} & 0.932 & 1.0 & \textbf{$\uparrow$0.068} \\

Flickr & 1.0 & 1.0 & \textbf{0} & 0.994 & 1.0 & \textbf{$\uparrow$0.006} & 0.996 & 1.0 & \textbf{$\uparrow$0.004} & 0.998 & 1.0 & \textbf{$\uparrow$0.002} \\
\hline
\end{tabularx}
\end{threeparttable}
\end{table*}

\begin{table*}[t!]
\centering
\normalsize
\tabulinesep=1.1mm
\caption{\mr{Comparison of TPRs ($\%$) under low FPRs ($1\%$) between Our Method and Baseline Membership Inference.}}
\label{tab:comparison_detection_baseline_TPR}
\begin{threeparttable}
\begin{tabularx}{\textwidth}{c|*{12}{>{\centering\arraybackslash}X}}
\hline
\multirow{2}{*}{} & \multicolumn{3}{c}{GCN} & \multicolumn{3}{c}{GraphSage} & \multicolumn{3}{c}{GAT} & \multicolumn{3}{c}{GIN} \\ \cline{2-13}
 & Baseline & Ours & $\Delta$ & Baseline & Ours & $\Delta$ & Baseline & Ours & $\Delta$ & Baseline & Ours & $\Delta$ \\ \hline
Cora & 0.257 & 0.945 & \textbf{$\uparrow$0.688} & 0.596 & 1.0 & \textbf{$\uparrow$0.404} & 0.202 & 0.953 & \textbf{$\uparrow$0.751} & 0.174 & 0.740 & \textbf{$\uparrow$0.566} \\

Citeseer & 0.832 & 1.0 & \textbf{$\uparrow$0.168} & 0.881 & 1.0 & \textbf{$\uparrow$0.119} & 0.897 & 1.0 & \textbf{$\uparrow$0.103} & 0.815 & 1.0 & \textbf{$\uparrow$0.185} \\

Pubmed & 0.618 & 1.0 & \textbf{$\uparrow$0.382} & 0.360 & 1.0 & \textbf{$\uparrow$0.640} & 0.717 & 1.0 & \textbf{$\uparrow$0.283} & 0.407 & 1.0 & \textbf{$\uparrow$0.593} \\

Flickr & 0.997 & 1.0 & \textbf{$\uparrow$0.003} & 0.966 & 1.0 & \textbf{$\uparrow$0.034} & 0.998 & 1.0 & \textbf{$\uparrow$0.002} & 0.998 & 1.0 & \textbf{$\uparrow$0.002} \\
\hline
\end{tabularx}
\end{threeparttable}
\end{table*}

\noindent \textbf{Evaluation Settings.} 
In the design of our graph data misuse mitigation framework, we have focused on four design requirements as shown in Sec. \ref{sec:design_req}. 

\noindent \textit{R1 - Misuse Detection Effectiveness}. To evaluate how our design can effectively detect misuse, 
we define a successful detection by the case when the data owner correctly identifies a node that is/is not used during the training. 
Since such misuse detection is a binary classification problem, we use the detection rate and calculate the corresponding \textit{AUC}. 
We evaluate the results among all nodes in both authorized and unauthorized graphs. 

\noindent \textit{R2 - Unlearning Effectiveness}. To evaluate how our design can effectively eliminate the impacts of unlearned nodes, we follow similar evaluation metrics to those of unlearning. 
Specifically, we use the \textit{attack successful rates} of the membership inference attack on both the unlearned and remaining nodes to evaluate the effectiveness of unlearning. 
The \textit{attack successful rate} of the membership inference attack in GNN is defined by the MIA attacker correctly identifying a node that is/is not used during training. 
We evaluate the results among all nodes in both authorized and unauthorized graphs. 

\noindent \textit{R3 - Data Privatization}. 
We have shown that our design satisfied R3 in our previous sections. 
That is, the data owner, model developer, and MLaaS server will not directly provide their data to other entities. 
Specifically, the data owner queries only isolated nodes without graph structure.
The model developer will also not provide his training graph structure to the MLaaS server during the unlearning process. 
The MLaaS server will also not provide direct access to the deployed GNN model to the data owner's request for misuse detection. 

\noindent \textit{R4 - GNN Model Agnostic}. 
We have shown that our design satisfies R4 since it does not require any specific GNN architectures as in our previous sections. 
In this section, we will also show that our design is generally suitable for different popular and common GNN architectures and datasets. 
In addition, we further evaluate the robustness of our misuse detection method by assuming that the model developer uses the graph denoising technique to update our radioactive graph data. 

\begin{figure*}[t!]
    \centering
    \begin{minipage}[htp]{0.24\linewidth}
        \centering
        \includegraphics[width=1\textwidth]{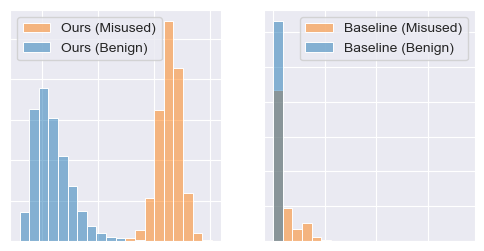}
        GCN-Cora
    \end{minipage} 
    \begin{minipage}[htp]{0.24\linewidth}
        \centering
        \includegraphics[width=1\textwidth]{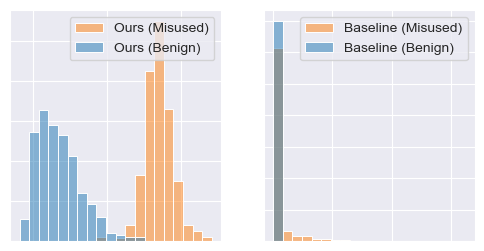}
        GCN-Citeseer
    \end{minipage}
    \begin{minipage}[htp]{0.24\linewidth}
        \centering
        \includegraphics[width=1\textwidth]{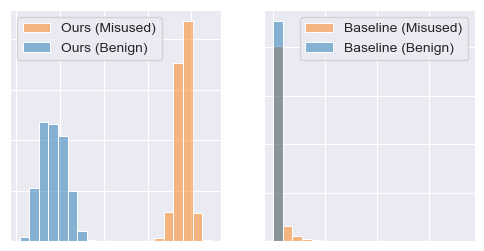}
        GCN-Pubmed
    \end{minipage}
    \begin{minipage}[htp]{0.24\linewidth}
        \centering
        \includegraphics[width=1\textwidth]{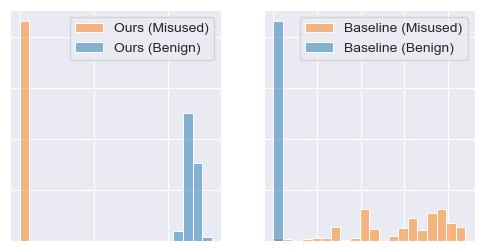}
        
        GCN-Flickr

    \end{minipage}
    \\
    \begin{minipage}[htp]{0.24\linewidth}
        \centering
        \includegraphics[width=1\textwidth]{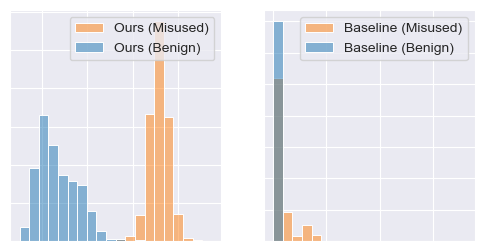}

        GraphSage-Cora

    \end{minipage}
    \begin{minipage}[htp]{0.24\linewidth}
        \centering
        \includegraphics[width=1\textwidth]{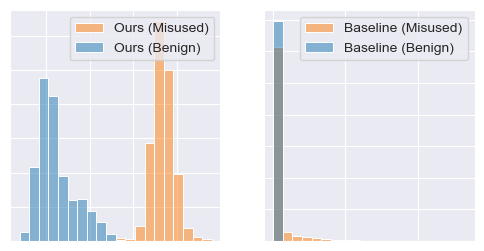}
        
        GraphSage-Citeseer

    \end{minipage}
    \begin{minipage}[htp]{0.24\linewidth}
        \centering
        \includegraphics[width=1\textwidth]{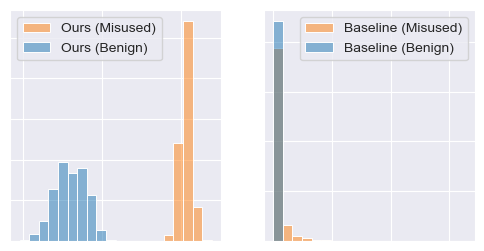}
        
        GraphSage-Pubmed

    \end{minipage}
    \begin{minipage}[htp]{0.24\linewidth}
        \centering
        \includegraphics[width=1\textwidth]{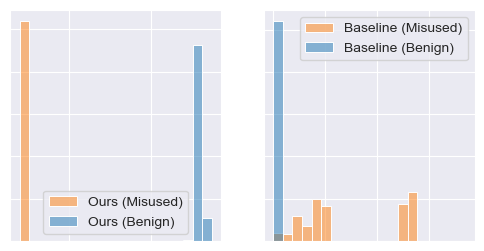}
        
        GraphSage-Flickr

    \end{minipage}
    \\
    \begin{minipage}[htp]{0.24\linewidth}
        \centering
        \includegraphics[width=1\textwidth]{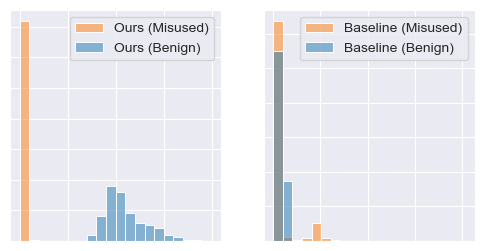}

        GAT-Cora

    \end{minipage}
    \begin{minipage}[htp]{0.24\linewidth}
        \centering
        \includegraphics[width=1\textwidth]{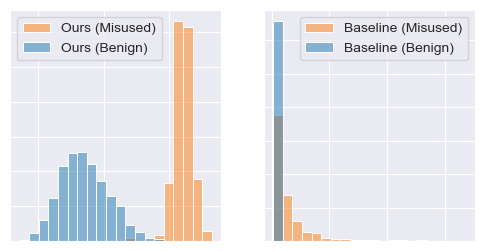}
        
        GAT-Citeseer

    \end{minipage}
    \begin{minipage}[htp]{0.24\linewidth}
        \centering
        \includegraphics[width=1\textwidth]{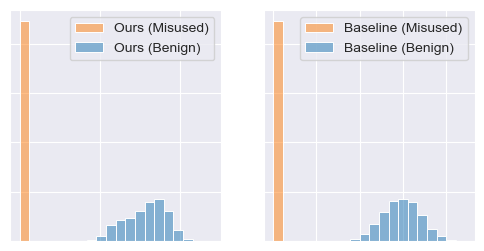}
        
        GAT-Pubmed

    \end{minipage}
    \begin{minipage}[htp]{0.24\linewidth}
        \centering
        \includegraphics[width=1\textwidth]{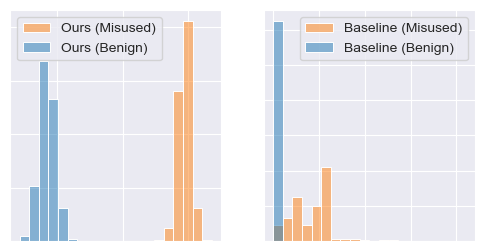}
        
        GAT-Flickr

    \end{minipage}
    \\
    \begin{minipage}[htp]{0.24\linewidth}
        \centering
        \includegraphics[width=1\textwidth]{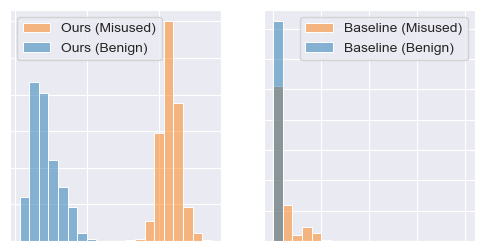}

        GIN-Cora

    \end{minipage}
    \begin{minipage}[htp]{0.24\linewidth}
        \centering
        \includegraphics[width=1\textwidth]{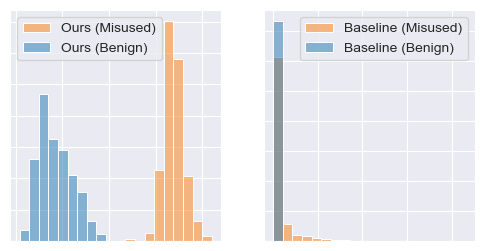}
        
        GIN-Citeseer

    \end{minipage}
    \begin{minipage}[htp]{0.24\linewidth}
        \centering
        \includegraphics[width=1\textwidth]{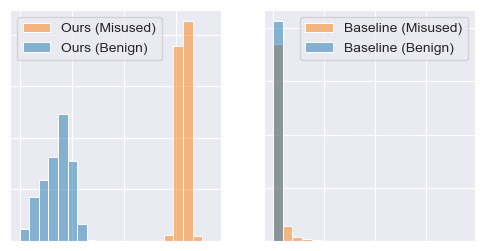}
        
        GIN-Pubmed

    \end{minipage}
    \begin{minipage}[htp]{0.24\linewidth}
        \centering
        \includegraphics[width=1\textwidth]{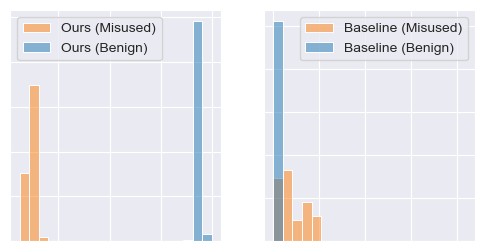}
        
        GIN-Flickr

    \end{minipage}
    \caption{Comparison of Output Distribution from Misused (in orange colors)/Benign (in blue colors) GNNs between Our method (left) and Baseline method (right). }
    \label{fig:distribution_comparison}
\end{figure*}

\noindent \textbf{Datasets.} Our design is evaluated on four widely recognized public graph datasets, namely, Cora~\cite{KipfW17}, Citeseer~\cite{KipfW17}, Pubmed~\cite{KipfW17}, and Flickr~\cite{ZengZSKP20}, all of which have been previously employed in GNN security analysis. A summary of the dataset statistics can be found in Tab.~\ref{tab:data_statistic}.
Cora, Citeseer, and Pubmed are citation networks where nodes signify publications, and node attributes denote the presence of specific keywords. 
An edge between nodes signifies a connection between two publications. 
Flickr is an image dataset, with nodes representing individual images and node attributes detailing the image profiles. 
An edge between nodes indicates common information between images, such as being originating from the same location or being submitted to the same gallery. 

\noindent \textbf{GNN Models.} 
We evaluate our design on four state-of-the-art GNN models: GCN~\cite{KipfW17}, GraphSage~\cite{HamiltonYL17}, Graph Attention Networks (GATs)~\cite{VelickovicCCRLB18}, and GIN~\cite{XuHLJ19}. 
The number of features in the hidden layer for all GNN models is 16. 
The activation function for the hidden layer is ReLU.
We use the Adam optimizer with a learning rate of 0.01 and training epochs of 300. 
The loss function of our model is the cross-entropy loss.

\noindent 
\mr{
\textbf{MLaaS Settings.}
For ethical considerations, we do not test our methods on a real MLaaS. Instead, to simulate the MLaaS setting, we restrict data access for each entity to resemble the environment inherent to MLaaS (see Tab.~\ref{tab:capabilities}). Furthermore, we emulate data transmission, ensuring that all transmitted data are the same as the communications in the actual MLaaS and that no private graph structure is transmitted through the system.
}



\subsection{Evaluating Proactive Misuse Detection - (R1, R4)}
This section evaluates the effectiveness of our proactive misuse detection module on four datasets and four GNN models by comparing membership inference results and distribution differences between member and non-member data.

\noindent \textbf{Effectiveness.} 
Tab.~\ref{tab:comparison_detection_baseline} shows the AUC of our detection method. 
It can be found that, for all four GNN models and datasets, our design can achieve almost 1 AUC. 
Namely, we can find a threshold to perfectly identify whether a node has been used during target model training. 
While our detection shows that it is effective among different datasets and GNN models, we further show that our design is general, which not only does not have an assumption on the training process and model architecture, but is also effective for different domain tasks and GNN models. 

\mr{
To demonstrate the effectiveness of our design in detecting graph misuse, we take into account the practical requirements for low false positive rates (FPRs) during detection. We evaluate TPRs (correctly detect misuse) with low FPRs (false alarms of detection). 
The results are presented in Tab.~\ref{tab:comparison_detection_baseline_TPR}. Across the four GNN models and datasets, our design consistently achieves much higher TPRs (i.e., almost $100\%$) than the baseline method with low FPRs, i.e., only $1\%$.
}

\noindent \textbf{Output Distribution Comparisons.} 
To further demonstrate the effectiveness of our radioactive graph construction in amplifying the performance disparity between benign and misuse GNN models, we compare the output distributions from both types of models in Fig.~\ref{fig:distribution_comparison}. 
Note that, in accordance with requirement R3, all these distributions are derived using $\hat{G}_{p}$, which safeguards the confidentiality of the graph structure. 
The results reveal that, without using our radioactive graph, the output distributions of benign and misuse models substantially overlap. 
This overlap arises because GNNs naturally learn from and potentially overfit to the graph structure. 
Consequently, the differences in output when querying $\hat{G}_{p}$, which does not contain graph structure information, are therefore negligible. 
However, with the introduction of our radioactive graph, the data-misused GNNs tend to learn more from the attributes, resulting in a marked distinction in performance compared to benign models. 
Based on the figures, there are only a few overlaps between the authorized and unauthorized nodes. 
Thus, it is convenient for the data owner to identify whether their graphs have been misused by the model developer. 

\noindent \textbf{Comparison with Baseline.} 
We consider the baseline following the strategies in~\cite{OlatunjiNK21}. 
Note that their method requires the data owner to query the exact authorized graph or have direct access to the suspect GNN model to obtain accurate predictions of the nodes. 
This method is more suitable to be applied locally, which does not satisfy R3 in our MLaaS settings. 
Tab.~\ref{tab:comparison_detection_baseline} compares our method with the baseline design. 
It can be found that, even with a stronger assumption on the knowledge of the inference graph data, our design can still achieve better performance than their method. 

\noindent \textbf{Robustness of Our Design.}
We further discuss the robustness of our design. 
In practice, model developers may use graph denoise methods to preprocess the training graph they gathered. 
This could happen occasionally since using these methods to improve the model performance is common. 
This could also be done by the advanced model developer who is aware of the proactive method that injects noise to the graph similar to us, and proposes to use the denoise method to remove these proactive perturbations.

Tab.~\ref{tab:detection_robustness} shows the AUC of the misuse detection with and without the denoising method in~\cite{Jin0LTWT20}. 
It could be found that, in all datasets and the GNN model, the AUC with the denoising method is almost the same as the AUC without the denoising method. 
Actually, this is because the graph denoise method commonly removes or adds the graph structure. 
However, our proactive design forces the GNN model to learn more from the attributes. 
Thus, common methods that denoising graphs by modifying the graph structure can not remove our injected proactive attributes. 
Namely, our design can be robust to these graph denoise methods. 

\begin{table*}[t!]
\centering
\normalsize
\tabulinesep=1.1mm
\caption{Comparison of AUC with/without Denoising Mechanism~\cite{Jin0LTWT20}.}
\label{tab:detection_robustness}
\begin{threeparttable}
\begin{tabularx}{\textwidth}{c|*{12}{>{\centering\arraybackslash}X}}
\hline
\multirow{2}{*}{} & \multicolumn{3}{c}{GCN} & \multicolumn{3}{c}{GraphSage} & \multicolumn{3}{c}{GAT} & \multicolumn{3}{c}{GIN} \\ \cline{2-13}
 & W & W/o & $\Delta$ & W & W/o & $\Delta$ & W & W/o & $\Delta$ & W & W/o & $\Delta$ \\ \hline
Cora & 1.0 & 0.999 & \textbf{$\downarrow$0.001} & 1.0 & 0.999 & \textbf{$\downarrow$0.001} & 1.0 & 1.0 & \textbf{0} & 1.0 & 1.0 & \textbf{0} \\

Citeseer & 1.0 & 0.999 & \textbf{$\downarrow$0.001} & 1.0 & 1.0 & \textbf{0} & 1.0 & 0.999 & \textbf{$\downarrow$0.001} & 1.0 & 1.0 & \textbf{0} \\

Pubmed & 1.0 & 1.0 & \textbf{0} & 1.0 & 1.0 & \textbf{0} & 1.0 & 1.0 & \textbf{0} & 1.0 & 1.0 & \textbf{0} \\

Flickr & 1.0 & 1.0 & \textbf{0} & 1.0 & 1.0 & \textbf{0} & 1.0 & 1.0 & \textbf{0} & 1.0 & 1.0 & \textbf{0} \\
\hline
\end{tabularx}
\end{threeparttable}
\end{table*}

\begin{table*}[t!]
    \centering
    \normalsize
    \tabulinesep=1.1mm
    \caption{Comparison of Model Accuracy (\% is omitted) between the Unlearned Model (denoted as $U$) and Retrained Model (denoted as $R$).}
    \label{tab:unlearning_utilities}
    \begin{threeparttable}
        \begin{tabularx}{\textwidth}{c|*{12}{>{\centering\arraybackslash}X}}
            \hline
            \multirow{2}{*}{} & \multicolumn{3}{c}{GCN} & \multicolumn{3}{c}{GraphSage} & \multicolumn{3}{c}{GAT} & \multicolumn{3}{c}{GIN} \\ \cline{2-13}
             & $R$ & $U$ & $\Delta$ & $R$ & $U$ & $\Delta$ & $R$ & $U$ & $\Delta$ & R & U & $\Delta$ \\ \hline
            Cora & 75.7 & 74.3 & \textbf{$\downarrow$ 1.2 } & 67.4 & 66.5 & \textbf{$\downarrow$ 0.9} & 83.1 & 81.5 & \textbf{$\downarrow$ 1.6} & 86.4 & 85.1 & \textbf{$\downarrow$ 1.3} \\
            Citeseer & 81.1 & 80.0 & \textbf{$\downarrow$ 1.1} & 70.0 & 68.7 & \textbf{$\downarrow$ 1.3} & 82.2 & 80.1 & \textbf{$\downarrow$ 2.1} & 79.5 & 78.9 & \textbf{$\downarrow$ 0.6} \\
            Pubmed & 81.8 & 79.8 & \textbf{$\downarrow$ 2.0} & 82.5 & 80.3 & \textbf{$\downarrow$ 2.2} & 83.6 & 81.3 & \textbf{$\downarrow$ 2.3} & 83.6 & 82.8 & \textbf{$\downarrow$ 0.8} \\
            \hline
        \end{tabularx}
    \end{threeparttable}
\end{table*}

\begin{table*}[t!]
\centering
\normalsize
\tabulinesep=1.1mm
\caption{Comparison of MIA Successful Rate(\% is omitted) of the Unlearned Nodes Before/After Unlearning.}
\label{tab:unlearning_MIA}
\begin{threeparttable}
\begin{tabularx}{\textwidth}{c|*{12}{>{\centering\arraybackslash}X}}
\hline
\multirow{2}{*}{} & \multicolumn{3}{c}{GCN} & \multicolumn{3}{c}{GraphSage} & \multicolumn{3}{c}{GAT} & \multicolumn{3}{c}{GIN} \\ \cline{2-13}
 & Before & After & $\Delta$ & Before & After & $\Delta$ & Before & After & $\Delta$ & Before & After & $\Delta$ \\ \hline
Cora & 86.9 & 51.8 & \textbf{$\downarrow$ 35.1} & 83.3 & 54.5 & \textbf{$\downarrow$ 28.8} & 85.6 & 47.5 & \textbf{$\downarrow$ 38.1} & 91.7 & 47.9 & \textbf{$\downarrow$ 43.8} \\

Citeseer & 91.3 & 68.7 & \textbf{$\downarrow$ 22.6} & 81.2 & 56.1 & \textbf{$\downarrow$ 25.1} & 61.4 & 60.3 & \textbf{$\downarrow$ 1.10} & 86.2 & 46.2 & \textbf{$\downarrow$ 40.0} \\

Pubmed & 93.6 & 49.2 & \textbf{$\downarrow$ 44.4} & 85.7 & 53.2 & \textbf{$\downarrow$ 32.5} & 82.4 & 49.7 & \textbf{$\downarrow$ 32.7} & 84.1 & 47.6 & \textbf{$\downarrow$ 36.5} \\
\hline
\end{tabularx}
\end{threeparttable}
\end{table*}

\begin{table*}[t!]
\centering
\normalsize
\tabulinesep=1.1mm
\caption{Comparison of Time Cost (in seconds) for Retraining the Model (denoted as $R$) and Our Unlearning Method.}
\label{tab:comparison_time}
\begin{threeparttable}
\begin{tabularx}{\textwidth}{c|*{12}{>{\centering\arraybackslash}X}}
\hline
\multirow{2}{*}{} & \multicolumn{3}{c}{GCN} & \multicolumn{3}{c}{GraphSage} & \multicolumn{3}{c}{GAT} & \multicolumn{3}{c}{GIN} \\ \cline{2-13}
 & $R$  & Ours & Times($\uparrow$) & $R$ & Ours & Times($\uparrow$) & $R$ & Ours & Times($\uparrow$) & $R$ & Ours & Times($\uparrow$)\\ \hline
Cora & 3.615 & 0.725 & $\approx$\textbf{4.99} & 4.188 & 0.643 & $\approx$\textbf{6.51} & 3.600 & 0.720 & $\approx$\textbf{5.0} & 4.26 & 1.225 & $\approx$\textbf{3.48} \\
Citeseer & 1.746 & 0.375 & $\approx$\textbf{4.66} & 2.023 & 0.333 & $\approx$\textbf{6.08} & 1.737 & 0.375 & $\approx$\textbf{4.63} & 2.058 & 0.613 & $\approx$\textbf{3.56} \\
Pubmed & 4.201 & 3.043 & $\approx$\textbf{1.38}  & 4.865 & 2.670 & $\approx$\textbf{1.82} & 4.190 & 3.017 & $\approx$\textbf{1.39} & 4.968 & 5.124 &  $\approx$\textbf{0.97} \\
\hline
\end{tabularx}
\end{threeparttable}
\end{table*}




\subsection{Evaluating Training-graph-free Unlearning - (R2, R4)}
In this section, we evaluate the effectiveness of our unlearning design on four datasets and four GNN models. 





\noindent \textbf{Effectiveness.} 
To validate the effectiveness of our unlearning design, we assess our unlearning method and perform experiments on three datasets and four GNN models. 
Tab.~\ref{tab:unlearning_MIA} illustrates the membership inference attack's accuracy of the different datasets and GNN models. 
It can be found that our design can significantly reduce the MIA attack success rate to about $0.5$ attack accuracy. 
Namely, it is a random guess of the membership of a target node (since MI is a binary classification problem). 
Meanwhile, the attack successful rate for MIAs targeting the model before unlearning is much higher (e.g., about $80\%$). 
Considering that MIA assesses how the model remembers training samples, our design can significantly eliminate the impact of these unlearning nodes. 

\noindent \textbf{Utility.} 
Additionally, we evaluate the utility of our unlearning method. Tab.~\ref{tab:unlearning_utilities} presents the accuracy of the GNN both prior to and after the unlearning process. 
It is evident that our approach incurs only a marginal decrease in the accuracy of the model, within $5\%$. 
This minor reduction in accuracy is reasonable, as the unlearning process involves the removal of some knowledge embedded in the unlearned samples. 
Consequently, trained GNNs possess less information from the training graph, which naturally leads to a slight decrease in the accuracy of the model.

\noindent \textbf{Time Cost.}
We compare the efficiency of our method with training from scratch that retrains the GNN in the remaining training set without unlearned samples. 
As shown in Tab.~\ref{tab:comparison_time}, we observe an efficiency improvement on all four GNN models,
We observed that the relative efficiency improvement of smaller datasets (Cora and Citeseer) is more than that of larger dataset (Pubmed). 
For instance, the unlearning time improvement is about 3-6× for both Cora, Citeseer dataset, and about 1-2× for the Pubmed dataset. 
This is because a larger amount of time is required to train a larger graph from scratch.
Although our method can significantly reduce the training efforts of the target model, it introduces new time costs for graph synthesis, which is also related to the size of the graph. 
However, we want to note that such computation cost is done by the MLaaS server which naturally contains more computation resources. 
Thus, our design can still be suitable for MLaaS settings.

\section{Discussion}
\label{sec:discussion}
\noindent 
\mr{
\textbf{Practicality in MLaaS Scenarios.}
Our design pipeline is well-suited for real-world MLaaS systems for two primary reasons:
1) Both proactive detection and training-graph unlearning only rely on using fundamental functions (e.g., training instance and prediction API) within MLaaS~\cite{dem108,awspdf,GoogleMlaas}. 
The detection performed by the data owner only queried the prediction APIs of the MLaaS server, which are basic functions provided by the MLaaS serving service~\cite{dem108,GoogleMlaas}. 
Similarly, the unlearning process executed by the server uses standard ML training and inference functions to generate unlearned graphs and unlearn the target model, which are available on general MLaaS platforms~\cite{dem108,awspdf,GoogleMLaaS_Training}.
2) Most of the computational overhead is borne by the server side. In our design, processes such as the training of the graph generation model or the fine-tuning of the target model are all handled by the MLaaS server, which possesses enough computational resources for these tasks~\cite{dem108,awspdf,GoogleMLaaS_Training}. 
}

\noindent 
\mr{
\textbf{Generalization to Other Tasks.}
%
Our design currently targets the node classification task in GNNs, but the underlying principles can be generalized to other tasks within GNNs. 
In nongraph-structured domains, similar strategies have been applied to image data and linear classifiers in~\cite{SablayrollesDSJ20} (we compare our design with theirs in Sec.~\ref{sec:proactiveMIA}). 
In other GNN tasks (e.g., graph classification and link prediction), our design can also be lent to them. 
Typically, the results for graph classification and link prediction are based on node embeddings, similar to those in node classification~\cite{GNNBook2022}. 
Given that our design aims to distinguish the output distributions derived from node embeddings between benign and data-misused GNNs, we can also vary distribution outcomes for both graph classification and link prediction using our strategies. 
We consider adaptation to other GNN tasks for future exploration. 
}

\noindent 
\mr{
\textbf{Generalization to Protecting the Node Attributes.}
Currently, our design focuses on protecting the graph structure. 
However, our approach can also be expanded to protect node attributes. 
Specifically, regarding misuse detection, our current strategy is to construct proactive graphs that remain effective for membership inference even when their graph structure undergoes perturbation (i.e., all edges are removed). 
To protect the node attribute, we can further update our current proactive graph construction algorithm to consider both the graph structure and the node attributes of the proactive graph to be perturbed (i.e., updating $X_p$ to $G_p=(A_p,X_p)$ in Equation~(\ref{eqt:PMIA_converted_opt})). 
Similarly, in the context of unlearning, our current methodology synthesizes only the unlearning training graph structure based on given node attributes. 
We can refine this by evolving our graph structure generation algorithm into one that simultaneously creates both the graph structure and node attributes. 
We envision this as a future step in our research.
}

\noindent 
\mr{
\textbf{Malicious Server.}
In this paper, our primary focus is on malicious model developers. 
Note that our design retains its capability to detect data misuse, even when the MLaaS server is malicious. 
Specifically, our detection method relies on querying the MLaaS server's prediction APIs (accessible to ordinary model users~\cite{dem108,awspdf,GoogleMlaas}). 
Once the server responds to these queries, the subsequent stages of the detection process are conducted locally on the data owner's side. 
Therefore, even in scenarios where the server rejects unlearning requests, our design remains effective in identifying data misuse.
}

\section{Related Work}
\label{sec:relatedwork}
\noindent \textbf{Membership Inference.} 
Existing studies demonstrates that machine learning models are vulnerable to attacks \cite{ZhangWY0WYP21, WuWYWRY23, QinHW23, ZhangYZP23, abs-2301-12951, ZhangWWYXPY23}.
For privacy attacks on GNNs \cite{abs-2205-07424}, according to the inference manner, existing membership inference attacks can be categorized into passive inference methods and active inference methods. Passive MIAs \cite{CarliniCN0TT22}, \cite{Yuan022}, \cite{0001MMBS22}, \cite{Choquette-ChooT21} aim to detect membership without elaborate active efforts (e.g., modifying data), while attackers in active MIAs  ~\cite{SablayrollesDSJ20}, \cite{TramerSJLJ0C22} will introduce elaborate efforts to facilitate detection.
However, it is not easy to directly adapt these studies on independent and identically distributed (IID) data to graph data, where connections between nodes break the IID assumption in these works. 

To this end, Olatunji et al.~\cite{OlatunjiNK21} first introduce MIAs in GNNs. Their approach was an extension of the general MIA methodologies in DNNs applied to GNNs. 
They employ a learning-based model to analyze the output confidence scores of target nodes' predictions and to ascertain their membership. 
On the other hand, Wu et al.~\cite{WuYPY21} applied MIAs to GNNs for graph classification. 
Their attack methods relied on the prediction probability vector and the metrics computed based on it, focusing primarily on graph-level GNN models.
In contrast, He et al.~\cite{abs-2102-05429} concentrated on MIAs against node-level classification GNNs. 
They considered the membership of a target node by evaluating the confidence scores for node prediction. 
This evaluation involved the target nodes and their immediate (0-hop) and secondary (2-hop) neighbors. 
Conti et al.~\cite{ContiLPX22} further studied label-only MIA against GNNs specifically for the node classification task. They hypothesized that attackers could obtain only node labels.

It is worth noting that all these MIAs require attackers to query the target model with the target graph data. 
This target graph comprises both node attributes and the graph structure, enabling a comprehensive analysis of the response, regardless of whether it is confidence scores or labels only. 
However, these attacks differ from our practical settings without directly querying the private target graph in MLaaS. 

\noindent \textbf{Machine Unlearning.} 
Based on the technology used for unlearning, current machine unlearning works include SISA-based methods \cite{BourtouleCCJTZL21} and others \cite{CaoY15}, \cite{WarneckePWR23}, \cite{ChundawatTMK23}, \cite{KimW22}, \cite{Zhang0ZCL22}. 
In SISA-based methods \cite{BourtouleCCJTZL21}, the training dataset is partitioned into multiple shards and then used to train multiple submodels individually, followed by integrating these submodels together to serve model users. 
Once unlearning requests are received, the model developer can only retrain one specific submodel to achieve fewer unlearning efforts \cite{abs-2209-02299}. 
Unlike SISA-based methods, other designs \cite{abs-2209-02299}, such as modifying model parameters with influence functions, can also be used in unlearning. More details can be found in recent machine unlearning surveys~\cite{abs-2209-02299, abs-2305-06360}. 

Since its inception by Cao et al.~\cite{CaoY15}, the concept of machine unlearning has generated a wealth of strategies for its implementation in GNNs. 
Among them, Chen et al.~\cite{Chen000H022} pioneered an approach named Graph Eraser, employing the Shard, Isolate, Sanitize, and Aggregate (SISA) method. 
They divided the training graph data into shards using graph partitioning, training each independently. Upon receiving an unlearning request, the model provider would retain the relevant shard model, trained on the subgraph that encompasses the unlearning data. 
Despite its effectiveness, it was dependent on an ensemble architecture and required retraining of the shard model using training graph data.

Chien et al.~\cite{chien2022certified} developed the first certified graph unlearning framework for GNNs, providing theoretical unlearning guarantees. 
However, their methodology was only applicable to linear GNNs. 
Similarly, Cheng et al.~\cite{cheng2023gnndelete} introduced GNNDelete, an unlearning strategy that involved adding extra weight matrices to negate the effect of unlearning data. 
Despite its ingenuity, the design required an additional block where the unlearning data was stored, reducing its practicality in the MLaaS setting.
Wu et al.~\cite{WuYQS0023} devised an unlearning strategy, defining an influence function known as the Graph Influence Function (GIF), which is considered an additional loss term for influenced neighbors, facilitating the unlearning of the graph structure. 
Moreover, Pan et al.~\cite{0003CM23} suggested an unlearning technique for GNNs that employed the Graph Scattering Transform, providing provable performance guarantees. 
However, both methods share a common requirement: precise unlearning samples must be utilized during the unlearning process to accurately eliminate the influence of specific data samples on the target unlearn model. 


\section{Conclusion}
\label{sec:conclusion}
In this paper, we introduce \textit{GraphGuard}, a pioneering integrated pipeline designed to address the graph data misuse issue in the context of GNNs deployed through MLaaS. 
Initially, we formalize the problem of graph data misuse concerning GNNs in the MLaaS setting and outline four critical requirements that reflect the practical considerations of MLaaS. 
Our proposed pipeline comprises two key components: a proactive graph misuse detection module, which adeptly detects graph misuse while respecting data privatization, and a training-graph-free unlearning module, which is capable of performing unlearning without access to training graph data and remains versatile across common GNN models. 
Through extensive experiments on four real-world graph datasets employing four state-of-the-art GNN models, we demonstrate the high effectiveness of GraphGuard.

\section*{Acknowledgements}
We would like to thank Neil Gong and Xu Yuan for their insightful discussions and feedback at the initial stage of this work and thank the anonymous shepherd and reviewers for their helpful comments and feedback. This work is supported in part by a Monash-Data61 Collaborative Research Project (CRP43) and Australian Research Council (ARC) DP240103068, FT210100097, and DP240101547. Minhui Xue, Xingliang Yuan and Shirui Pan are also supported by CSIRO – National Science Foundation (US) AI Research Collaboration Program. 

\bibliographystyle{IEEEtranS}
\balance
\bibliography{reference}

\section*{Appendix}
\subsection{Radioactive Graph Construction}
\label{app:radioactive_construction_proof}
\noindent \textbf{Function-similar MI Model - Substituting $\mathcal{A}$.}
Since acquiring a well-trained MI model $\mathcal{A}$ may also be challenging for the data owner, our design employs a simplified but practical inference model to substitute the ideal $\mathcal{A}$.
Note that the primary function of $\mathcal{A}$, whose value lies in the range $[0,1]$, is to capture the overfitting of the target models, i.e., to ideally output $1$ on members (i.e., $G_m$, training/overfitted samples)  and 0 on non-members (i.e., $G_n$, testing/under fitted samples) of the target GNN $f_{\theta^{*}}$.
Instead of training $\mathcal{A}$ with the shadow dataset and model, as shown by the practical MI in Sec. \ref{sec:MIA_background}, we propose to substitute the well-trained $\mathcal{A}$ with a function-similar while practical indexing function $\hat{\mathcal{A}}$, which is underpinned by the intuition that the learning of $\mathcal{A}$ is to distinguish the difference of confidence scores between members and non-members of GNNs.
Formally, $\hat{\mathcal{A}}$ is defined as
\begin{equation}
\label{eq:mi_substitution}
    \begin{aligned}
        \hat{\mathcal{A}}(f_{\theta^{*}}(G)_{v_i}) = f_{\theta^{*}}(G)_{v_i}[y_i],
    \end{aligned}
\end{equation}
where $f_{\theta^{*}}(G)_{v_i}$ denotes the predicted labeling result on the node $v_i$ from the input graph $G$,
$[y_i]$ indicates the indexing function which selects the $y_i$-th element in vector $f_{\theta^{*}}(G)_{v_i}$, and $y_i$ represents the ground-truth label of $v_i$. Note that, given Equation (\ref{eq:mi_substitution}), $\hat{\mathcal{A}}$ denotes the prediction confidence score of the nodes in the query graph.

Incorporating $\hat{G}_p$ and $\hat{\mathcal{A}}$, the basic graph data misuse detection strategy in MLaaS is defined as follows:
Given a suspect target GNN $f_{\theta}^{*}$ trained on $G_m$, which may have unauthorizedly used the target graph $G_p = (A_p, X_p)$ for training (i.e., $G_m \cap G_p \neq \emptyset$), the data owner of $G_p$ can detect misuse by querying this target GNN $f_{\theta^{*}}$ with graph $\hat{G}_p = (\hat{A}_p, X_p)$, followed by calculating a membership confidence score $\mathtt{mcs}$ for $G_p$:
\begin{equation}
        \mathtt{mcs} = \frac{1}{|V_{\hat{G}_p}|}\sum_{i=1}^{|V_{\hat{G}_p}|}\hat{\mathcal{A}}(f_{\theta^{*}}(\hat{G}_p)_{v_i}),
    \label{eqt:pra_MIA}
\end{equation}
where $|V_{\hat{G}_p}|$ indicates the node number of $\hat{G}_p$. 
When $\mathtt{msc}$ is close to 1, it is more likely that $G_p$ was used during the training phase, while a $\mathtt{msc}$ value closer to 0 indicates that it is less probable that $G_p$ was used during the training of $f_{\theta^{*}}$. 

\noindent \textbf{Objective Function Convertion.} 
In this paper, from the view of the data owner, a desired $G_p$ should satisfy $\mathcal{A}$ is expected to satisfy $\mathcal{A}(f_{\theta^{*}}(G_p))=1$ when $\theta^{*}$ is influenced by $G_p$ during the training of $f$, otherwise $\mathcal{A}(f_{\theta^{*}}(G_p))=0$. 
Similarly, $G_p$ and $\mathcal{A}$
are respectively substituted by $\hat{G}_p$ and $\hat{\mathcal{A}}$ to avoid exposing the sensitive graph structure
and meet the practical capability of the data owner. 
Therefore, we propose that the proactive graph $G_p$ can be obtained by solving the following optimization to help identify if $G_p$ is used in the training of the target GNN $f_{\theta^{*}}$:
\begin{equation}
\label{eq:rgraph_op_app}
\begin{aligned}
\min_{G_p}\ & 1- \hat{\mathcal{A}}(f_{\theta^{*}_{1}}(\hat{G}_p)) + \hat{\mathcal{A}}(f_{\theta^{*}_{0}}(\hat{G}_p)), \\
s.t. \ & \theta^{*}_{1} =\arg \min_{\theta} L(f_{\theta}(G_m^{1})),\\
& \theta^{*}_{0} =\arg \min_{\theta} L(f_{\theta}({G}_m^{0})),\\
\end{aligned} 
\end{equation}
where $G_m^{1}$ and $G_m^{0}$ represent two different versions of $G_m$, namely, $G_m^{1}$ indicates the $G_m$ which (partially) includes  $G_p$ (i.e., $G_p \cap G_m^{1} \neq \emptyset$) while $G_m^{0}$ denotes the $G_m$ which excludes $G_p$ (i.e., $G_p \cap {G}_m^{0} = \emptyset$). 
Note that the first element $1- \hat{\mathcal{A}}(f_{\theta^{*}_{1}}(\hat{G}_p))$ targets the case ${G}_m^{1}$, while the second item $\hat{\mathcal{A}}(f_{\theta^{*}_{0}}(\hat{G}_p))$ is designed for the case ${G}_m^{0}$.

\noindent \textbf{Optimization Solving.} 
Obtaining $\hat{G}_p$ from (\ref{eq:rgraph_op_app}) is non-trivial due to two reasons. 
\textbf{(a)} \textit{Uncertain $f_{\theta^{*}}(\hat{G}_p)$}. Given that the data owner can only query $f_{\theta^{*}}$ in the context of MLaaS, the data owner cannot determine the source of the querying results (i.e., $f_{\theta^{*}_{1}}(\hat{G}_p)$ or $f_{\theta^{*}_{0}}(\hat{G}_p)$), resulting in a dilemma for the data owner when selecting the optimization goal (i.e., either $\min_{\hat{G}_p}1- \hat{\mathcal{A}}(f_{\theta^{*}_{1}}(\hat{G}_p))$ or $\min_{\hat{G}_p}\hat{\mathcal{A}}(f_{\theta^{*}_{0}}(\hat{G}_p))$).
\textbf{(b)} \textit{Either-or Optimization}. Assuming that the source of query results $f_{\theta^{*}}(\hat{G}_p)$ has been determined, only one item (i.e., $1- \hat{\mathcal{A}}(f_{\theta^{*}_{1}}(\hat{G}_p))$ or $\hat{\mathcal{A}}(f_{\theta^{*}_{0}}(\hat{G}_p))$) can be involved in solving the whole optimization (\ref{eq:rgraph_op_app}), which potentially contributes to sub-optimal $\hat{G}_p$.

To this end, we propose to devise customized solutions for different items in optimization \ref{eq:rgraph_op_app}, and then combine these solutions together with a novel \textit{feature-loss-max} principle.
\noindent
\textbf{(1) Customized Solutions.} For items $1- \hat{\mathcal{A}}(f_{\theta^{*}_{1}}(\hat{G}_p))$ and $\hat{\mathcal{A}}(f_{\theta^{*}_{0}}(\hat{G}_p))$, the customized solutions are listed below.

\noindent
\textbf{Case-A:} $1- \hat{\mathcal{A}}(f_{\theta^{*}_{1}}(\hat{G}_p))$.
In addition to the above two reasons, another challenge in solving $\min_{G_p}1- \hat{\mathcal{A}}(f_{\theta^{*}_{1}}(\hat{G}_p))$ is the interdependence between $G_p$ and $\theta^{*}_{1}$. On the one hand, $\theta^{*}_{1}$ is derived from $\theta^{*}_{1} =\arg \min_{\theta} L(f_{\theta}(G_m^{1}))$, where $G_p \cap G_m^{1} \neq \emptyset$; on the other hand, the calculation of $\hat{G}_p$ is based on $\theta^{*}_{1}$. This interdependence further makes the optimization difficult to solve.

To this end, we propose to regard $\min_{\hat{G}_p}1- \hat{\mathcal{A}}(f_{\theta^{*}_{1}}(\hat{G}_p))$ as a clean-label poisoning attack, wherein the attacker adds elaborate poisoning examples to the training dataset to manipulate the behavior of the poisoned target model at the test time.
In the context of this paper, $\mathcal{A}(f_{\theta^{*}_{1}}(\cdot))$ represents the poisoned model, 
and $G_p$ is the poisoning graph data, whose goal of $G_p$ is to make the poisoned model $\mathcal{A}(f_{\theta^{*}_{1}}(\cdot))$ classifies $\hat{G}_p$ as a specific target label 1 (member).
According to the existing literature, the \textit{feature collision} method~\cite{ShafahiHNSSDG18} is an effective solution for this clean-label poisoning attack, whose attack paradigm can be instantiated considering the context of this paper as 
\begin{equation}\min_{G_p}\ D(f_{\theta^{*}_{1}}(G_p),f_{\theta^{*}_{1}}(\hat{G}_p)),
\label{eqt:feat_prac_d}
\end{equation}
where $G_p= (A_p,X_p)$ is the radioactive graph, $\hat{G}_p = (\hat{A}_p,X_p)$ represents the querying version of $G_p$ with isolated node sets (i.e., $\hat{A}_p = I_{|V_{G_p}|}$), and $D(\cdot,\cdot)$ represent the distance function.
Note that the feature collision \cite{ShafahiHNSSDG18} has shown transferability for different classifiers (i.e., $\mathcal{A}$ in our case)~\cite{ShafahiHNSSDG18}, which underpins neglecting the impact of $\mathcal{A}$ in Equation (\ref{eqt:feat_prac_d}). 

\noindent
\textbf{Case-B:} ${\hat{\mathcal{A}}(f_{\theta^{*}_{0}}(\hat{G}_p))}$.
Similarly, we propose to consider $\min_{G_p}\ \hat{\mathcal{A}}(f_{\theta^{*}_{0}}(\hat{G}_p))$ as an evasion attack, where the original graph data of data owner is perturbed as $G_p$ to make the target model $\hat{\mathcal{A}}(f_{\theta^{*}_{0}}(\cdot))$ predict 0 in $\hat{G}_p$. Note that $\hat{\mathcal{A}}(f_{\theta}(G)_{v_i}) = f_{\theta}(G)_{v_i}[y_i]$ as shown in Equation (\ref{eq:mi_substitution}), thus $\min_{G_p}\ \hat{\mathcal{A}}(f_{\theta^{*}_{0}}(\hat{G}_p))$ can be transferred as
\begin{equation}
\label{eqt:feat_prac_d:second}
        \max_{G_p} L(f_{\theta^{*}_{0}}(\hat{G}_p),Y_p)
\end{equation}
where $Y_p$ represents the ground-truth labels of nodes in $\hat{G}_p$, and $L(\cdot,\cdot)$ indicates the loss function.

\noindent \textbf{(2) Solution Integration.}
To combine the above optimization solutions together, we propose a straightforward but effective approach called \textit{feature-loss-max}, which can unify Equations (\ref{eqt:feat_prac_d}) and (\ref{eqt:feat_prac_d:second}) in the same view and then make solving the whole optimization (\ref{eq:rgraph_op_app}) feasible.
The intuition behind our \textit{feature-loss-max} is that a machine learning model will preferentially fit training samples with larger prediction loss when all samples are treated equally. 
This intuition indicates that improving the prediction loss of $f_{\theta^{*}}$ in $G_p$ contributes to improving the influence of $G_p$ on $\theta^{*}$, which helps improve the difference between $\theta^{*}_{1}$ and $\theta^{*}_{0}$ and then facilitates the distinguishing of $f_{\theta^{*}_{1}}$ and $f_{\theta^{*}_{0}}$.
Next, we will introduce how to project Equations (\ref{eqt:feat_prac_d}) and (\ref{eqt:feat_prac_d:second}) into the view of feature (i.e., $X_p$ in $G_p$) optimization.

\noindent
\textbf{Case-A:} Optimization (\ref{eqt:feat_prac_d}) suggests that the constructed $G_p$ will make the target model $f_{\theta^{*}_{1}}$ focus on the node features but not on the graph structure, since $G_p= (A_p,X_p)$ and $\hat{G}_p = (I_{|V_{G_p}|},X_p)$. Therefore, we propose to mainly modify $X_p$ when optimizing (\ref{eqt:feat_prac_d}), i.e., $\min_{X_p}\ D(f_{\theta^{*}_{1}}(G_p),f_{\theta^{*}_{1}}(\hat{G}_p))$. 
We also note that: 
\textbf{(a)} According to the $(D_1,D_2)$-Lipschitz property \cite{abs-2205-07424} of the GNN models, we obtain the following. 
\begin{equation*}
    D_{1}(f_{\theta^{*}_{1}}(G_p),f_{\theta^{*}_{1}}(\hat{G}_p)) \leq L_{f_{\theta^{*}_{1}}^{c}} D_{2}(f_{\theta^{*}_{1}}^{e}(G_p),f_{\theta^{*}_{1}}^{e}(\hat{G}_p)), 
\end{equation*}
where $L_{f_{\theta^{*}_{1}}^{c}}$ indicates a constant scalar decided by $f_{\theta^{*}_{1}}^{c}$, $D_1$ and $D_2$ represents two distance functions, $f_{\theta^{*}_{1}}^{c}$ and $f_{\theta^{*}_{1}}^{e}$ indicate the classifier and encoder in the target GNN and $f_{\theta^{*}_{1}} = f_{\theta^{*}_{1}}^{c} \circ f_{\theta^{*}_{1}}^{e}$, symbol $\circ$ represents the function composition operation. 
Therefore, we conclude that reducing the difference between $f_{\theta^{*}_{1}}^{e}(G_p)$ and $f_{\theta^{*}_{1}}^{e}(\hat{G}_p)$ is equivalent to solving the optimization (\ref{eqt:feat_prac_d}). 
\textbf{(b)} To avoid exposing the graph structure, we cannot deliver $G_p$ to $f_{\theta^{*}_{1}}^{e}$. Thus, we propose to employ a pre-trained encoder $f_{\theta^{p}}^{e}$ to perform as $f_{\theta^{*}_{1}}^{e}$.
In summary, our final optimization for the item $1- \hat{\mathcal{A}}(f_{\theta^{*}_{1}}(\hat{G}_p))$ in Equation (\ref{eq:rgraph_op_app}) is as follows. 
\begin{equation}
\label{eqt:feat_prac_d:final}
    \min_{X_p}\ ||f_{\theta^{p}}^{e}(G_p)-f_{\theta^{p}}^{e}(\hat{G}_p)||_2
\end{equation}
where $||\cdot||_2$ indicates the the $\ell_2$-norm operation.

\noindent
\textbf{Case-B:} 
Note that the nodes in $\hat{G}_p$ are isolated (i.e., $\hat{A}_p = I_{|V_{G_p}|}$), only the shared node features between $\hat{G}_p$ and $G_p$ are optimized when solving Equation (\ref{eqt:feat_prac_d:second}). Simultaneously, a surrogate model is needed when the data owner cannot ensure whether the target GNN $f_{\theta^{*}}$ is $f_{\theta^{*}_{0}}$. Thus, our optimization for the item ${\hat{\mathcal{A}}(f_{\theta^{*}_{0}}(\hat{G}_p))}$ in Equation (\ref{eq:rgraph_op_app}) is 
\begin{equation}
\label{eqt:feat_prac_d:second:final}
        \max_{X_p} L(f_{\theta^{p}}(\hat{G}_p),Y_p),
\end{equation}
where $f_{\theta^{p}} = f_{\theta^{p}}^{c} \circ f_{\theta^{p}}^{e}$ is a pre-trained surrogate model, which is consistent with prior works on graph unlearning~\cite{Chen000H022}.
Note that $f_{\theta^{p}}$ can be any publicly available GNN model that performs similar tasks or a handy GNN model trained on small-scale graph data. 

\noindent
\textbf{Integration (A + B):}
Integrating the Equations (\ref{eqt:feat_prac_d:final}) and (\ref{eqt:feat_prac_d:second:final}) together, the radioactive graph in this paper is obtained by solving
\begin{equation}
        \min_{X_p}\ ||f_{\theta^{p}}^{e}(G_p)-f_{\theta^{p}}^{e}(\hat{G}_p)||_2 - L(f_{\theta^{p}}(\hat{G}_p),Y_p).
\end{equation}
Intuitively, the constructed $G_p= (A_p,X_p)$ will make the GNN trained in it (1) learn less about confidential $A_p$ to protect the privatization of the data owner, and (2) produce a greater training loss in $G_p$, which increases its influence on the target GNNs and then facilitates the misuse detection. 

\clearpage




\section*{Artifact Appendix}

\textbf{
This artifact is based on PyTorch and the deep graph library (DGL), and does not require GPU support. 
We implemented the proposed mitigation pipeline against data misuse in GNNs, including Proactive Misuse Detection and Training-graph-free Unlearning of our paper ``GraphGuard: Detecting and Counteracting Training Data Misuse in Graph Neural Networks''. 
The artifact can reproduce our major experimental results (AUC of the data misuse detection, visualization of the diverse output confidence scores distributions between benign/misused GNNs, accuracy and MIA attack success rate of GNNs after unlearning) reported in the main body of the paper. 
Our source code is available at
\url{https://github.com/GraphGuard/GraphGuard-Proactive/README.md}, with
a detailed guide at 
\url{https://github.com/GraphGuard/GraphGuard-Proactive/AE/reproduce.md}
. 
We also uploaded the artifact to Zendo, and the DOI is \textit{10.5281/zenodo.10201956}.
}

\subsection{Description \& Requirements}

\subsubsection{Security, privacy, and ethical concerns}

None. All data misuse attacks against GNNs in our artifact are
conducted on the simulation level, therefore do not lead to
any damage in the real world.

\subsubsection{How to access}
Our artifact source code is hosted in a GitHub repository, available through 
\url{https://anonymous.4open.science/r/Proactive-MIA-Unlearning-BBC2/README.md}
. 
We will make our source code public in the camera-ready version.

\subsubsection{Hardware dependencies}
This artifact requires a Linux server and does not require GPU support. 

\subsubsection{Software dependencies} 
This artifact relies on multiple existing Python packages, including Python, PyTorch, DGL, and so on (details in \texttt{Init environment} section in \texttt{README.md}). 
Our guide includes all the details to set up the required software dependencies. 

\subsubsection{Benchmarks} 
Our experiments with this artifact are reported on 4 benchmark datasets: Cora, Citeseer, Pubmed (three citation networks commonly used for evaluating node classification tasks in GNNs), Flickr (an image dataset with profiles, where a node represents an image, and an edge represents two images have common information). 
All of them are integrated into DGL and automatically downloaded and setup in our artifact.

\subsection{Artifact Installation \& Configuration}

Our documentation contains a detailed guide for installing our artifact and required environments. 
Briefly, the installation procedure is as follows.

\begin{itemize}
    \item [1] Get the artifact from \url{https://github.com/GraphGuard/GraphGuard-Proactive}.
    \item [2] Install Conda following the instructions in \url{https://conda.io/projects/conda/en/latest/user-guide/install/index.html} to install required environments. 
    \item [3] Install Pytorach and other necessary Python packages using instructions in \texttt{Init environment} section in \texttt{README.md}. 
\end{itemize}

\subsection{Major Claims}

\begin{itemize}
    \item (C1): Proactive Misuse Detection in our proposed design \textit{GraphGuard} can strengthen the ability to discern benign and misused model performance. 
    This is proven by the experiment (E2) whose results are illustrated in Table IV. 
    It also achieves better detection performance (higher AUC) compared to existing studies. 
    This is proven by the experiment (E1) whose results are illustrated in Figure 4. 
    \item (C2): Training-graph-free Unlearning in \textit{GraphGuard} can mitigate data misuse in GNNs and decrease the MIA successful rate of unlearned nodes before/after unlearning. 
    This is proven by the experiment (E3) whose results are illustrated in Table VII. 
    It will also not lead to significant deduction of the model accuracy. 
    These are proven by the experiment (E4) whose results are illustrated in Table VI. 
\end{itemize}

\subsection{Evaluation}






\subsubsection{Experiment (E1)}

[Effectiveness of Proactive Misuse Detection]: Experiment (E1) evaluates the effectiveness of our proposed Proactive Misused Detection and compares it to the baseline membership inference attacks (MIAs) across 4 different models and datasets. 
It calculates the AUC of both our detection and the baseline detection via MIAs, thus proving our first claim (C1). 
Experiment (E1) corresponds to our reported results in Table IV. 

\textit{[How to]} All the preparation steps have been stated
in Artifact Appendix B. We provide all the necessary commands to reproduce our results of (E1) in \texttt{AE/reproduce.md\#Results-for-E1}. 

\textit{[Results]}
The AUC of our proposed detection is almost $1$, therefore proving that our design is highly effective in detecting data misuse. And the AUC of the baseline MIA is less than our proposed design. 

\subsubsection{Experiment (E2)}

[Output Distribution Comparisons]: Experiment (E2) evaluates the ability of our proposed Proactive Misused Detection about how it can discern the output distribution of benign and misused models. 
It compares the output distribution across 4 different models and datasets, therefore proving our first claim (C1). 
Experiment (E2) corresponds to our reported results in Figure 4. 

\textit{[How to]} All the preparation steps have been stated
in Artifact Appendix B. We provide all the necessary commands to reproduce our results of (E2) in \texttt{AE/reproduce.md\#Results-for-E2}. 

\textit{[Results]}
The difference between the output distribution of benign/misused models with our design is significant, while there is an overlap between the output distribution of benign/misused models without our design. 

\subsubsection{Experiment (E3)}

[Effectiveness of Training-graph-free Unlearning] Experiment (E3) evaluates the effectiveness of our proposed Training-Graph-Free Unlearning. 
It calculates the MIAs of the data-misused model with/without using unlearning, which proves our second claim (C2). 
The results correspond to the results reported in Table VII. 

\textit{[How to]} All the preparation steps have been stated
in Artifact Appendix B. We provide all the necessary commands to reproduce our results of (E3) in \texttt{AE/reproduce.md\#Results-for-E3}. 

\textit{[Results]}
The attack success rate of the MIAs is significantly reduced after using our unlearning method.

\subsubsection{Experiment (E4)}

[Utility of Training-graph-free Unlearning] Experiment (E4) evaluates the utility of our proposed Training-Graph-Free Unlearning. 
It also compares the accuracy of the unlearned model and the retrained model as a baseline. 
The results correspond to the results reported in Table VI. 

\textit{[How to]} All the preparation steps have been stated
in Artifact Appendix B. We provide all the commands necessary to reproduce our results of (E4) in \texttt{AE/reproduce.md\#Results-for-E4}. 

\textit{[Results]}
The accuracy of our unlearned model is only slightly lower than the retraining baseline (less than $5\%$).


\end{document}